\documentclass[letterpaper]{article}
\usepackage[ccby]{aaai2027_arxiv}
\usepackage[hyphens]{url}
\usepackage{graphicx}
\usepackage{natbib}
\usepackage{caption}
\usepackage[table]{xcolor}
\usepackage{amsmath,amssymb,amsfonts,amsthm}
\usepackage{mathtools}
\newtheorem{proposition}{Proposition}

\usepackage{booktabs}
\usepackage{multirow}
\usepackage{algorithm}
\usepackage{algpseudocode}
\usepackage{array}
\usepackage{enumitem}
\usepackage[colorlinks=true, allcolors=blue]{hyperref}
\hypersetup{
  pdftitle={ATOD: Annealed Turn-Aware On-Policy Distillation for Multi-Turn Agentic Tasks},
  pdfauthor={Qitai Tan, Zefang Zong, Mo Li, Yipeng Shi, Yang Li, Peng Chen},
  pdfsubject={arXiv preprint},
}
\usepackage[capitalise,noabbrev]{cleveref}

\newcommand{\method}{ATOD}

\newcommand{\student}{\pi_{\theta}}
\newcommand{\oldstudent}{\pi_{\theta_{\mathrm{old}}}}
\newcommand{\teacher}{\pi_{\mathrm{T}}}
\newcommand{\loss}{\mathcal{L}}
\newcommand{\traj}{\tau}
\newcommand{\stepset}{\mathcal{I}}

\newcommand{\tdur}{\ensuremath{\mathrm{T\text{-}DUR}}}
\newcommand{\grpo}{\mathrm{GRPO}}
\newcommand{\opd}{\mathrm{OPD}}

\definecolor{casebar}{HTML}{4D4D4D}
\definecolor{casegreen}{HTML}{D9F7D1}
\definecolor{caseyellow}{HTML}{FFF6C7}
\definecolor{caseorange}{HTML}{FFE2BE}
\definecolor{casered}{HTML}{F7C9C9}
\newcolumntype{L}[1]{>{\raggedright\arraybackslash}p{#1}}
\newcolumntype{C}[1]{>{\centering\arraybackslash}p{#1}}
\newcommand{\casew}[2]{\cellcolor{#1}\textbf{#2}}

\title{ATOD: Annealed Turn-Aware On-Policy Distillation for Multi-Turn Agentic Tasks}
\author{
Qitai Tan\textsuperscript{1,2} \quad
Zefang Zong\textsuperscript{1} \quad
Mo Li\textsuperscript{2} \quad
Yipeng Shi\textsuperscript{1,3} \quad
Yang Li\textsuperscript{1} \quad
Peng Chen\textsuperscript{1}
}
\affiliations{
\textsuperscript{1}Tencent Inc. \qquad
\textsuperscript{2}Tsinghua University \qquad
\textsuperscript{3}Peking University \\[0.6em]
\texttt{tqt24@mails.tsinghua.edu.cn} \\
\texttt{\{willzong, thomasyngli, pengchen\}@tencent.com}
}

\begin{document}
\maketitle

\begin{abstract}
Training small language-model agents for long-horizon interactive tasks requires both fast imitation and reward-driven improvement. On-policy distillation (OPD) provides dense teacher guidance and typically improves rapidly in the early stage, but its gains saturate once the student approaches the teacher, limiting the final performance ceiling. Reinforcement learning (RL) directly optimizes environment rewards and encourages exploratory improvement toward a higher reward-defined ceiling, but sparse and delayed feedback makes early-stage learning much less efficient than OPD. In this paper, we propose \textbf{\method{}} (\textbf{A}nnealed \textbf{T}urn-aware \textbf{O}n-policy \textbf{D}istillation), a hybrid online distillation algorithm that explicitly exploits this complementarity. \textbf{(1)} \method{} combines OPD and RL in a single hybrid objective with a \textbf{smoothly annealed weighting schedule}: teacher guidance is emphasized during bootstrapping, while the reward coefficient is progressively increased to support exploration beyond imitation. \textbf{(2)} \method{} introduces \textbf{Turn-level Disagreement-Uncertainty Reweighting} (\tdur{}), which softly gates the distillation signal to prioritize turns with high disagreement or uncertainty in long trajectories. Experiments on ALFWorld, WebShop, and Search-QA show that \method{} attains the highest average success rate at all three student sizes and the best per-task result in eight of the nine benchmark--student combinations: \method{} improves average success rate by \textbf{4.16} points over OPD and \textbf{23.62} points over GRPO, and surpasses the corresponding teacher models by \textbf{2.16} points on average. Code is available at \url{https://github.com/TanQitai/ATOD}.
\end{abstract}

% Figure 1 (benchmark bar) removed from main text; see appendix or supplementary material.

\section{Introduction}
\label{sec:introduction}

\begin{figure}[t]
\centering
\includegraphics[width=\columnwidth]{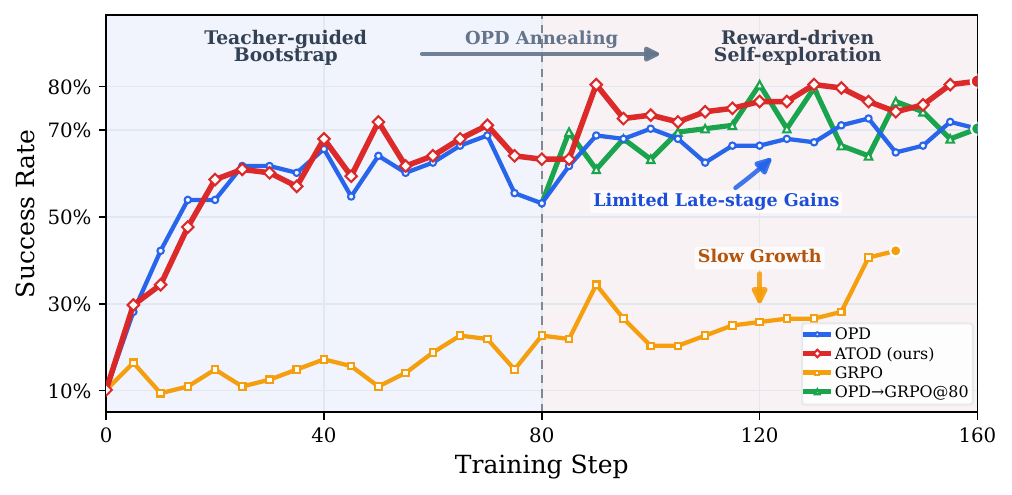}
\caption{Training dynamics of the 1.7B student on ALFWorld under \method{}, OPD, GRPO, and OPD$\rightarrow$GRPO@80. OPD$\rightarrow$GRPO@80 follows OPD for the first 80 steps and then switches to GRPO for the remaining 80 steps.}
\label{fig:annealing-intuition}
\end{figure}

\begin{figure*}[t]
\centering
\includegraphics[width=\textwidth]{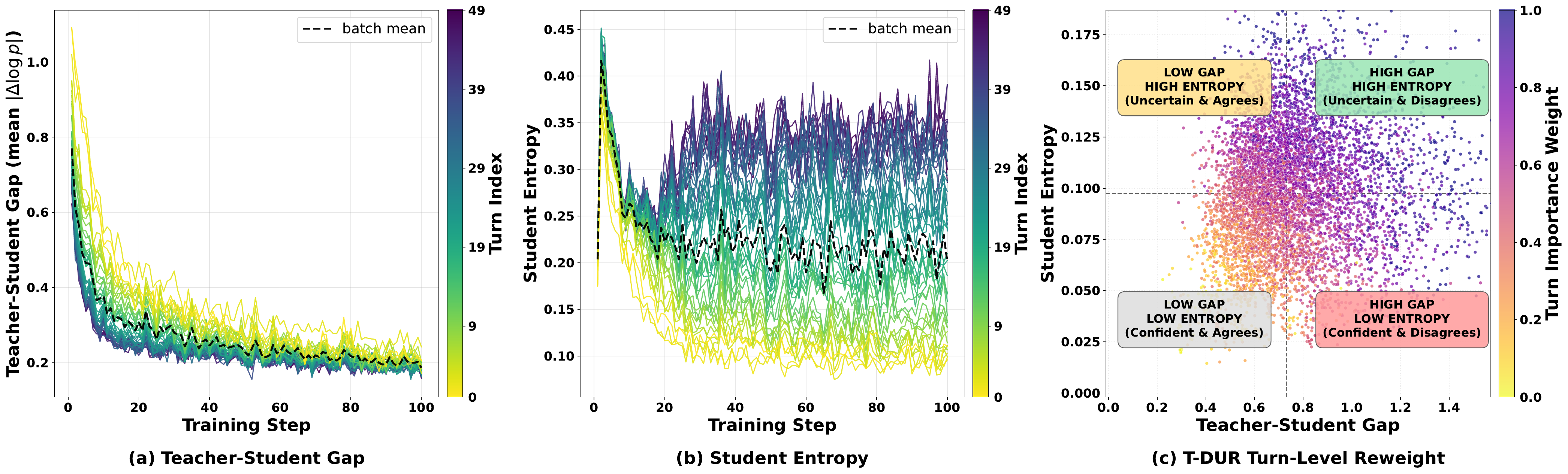}
\caption{Turn-level diagnostics for the 1.7B student under vanilla OPD on ALFWorld (darker curves denote deeper turns). (a) Shallow turns consistently exhibit a larger teacher--student log-probability gap than deep turns. (b) As training progresses, uncertainty decreases on shallow turns but remains elevated on deeper turns. Together, these observations show that a small observed gap does not necessarily imply low learning need. (c) T-DUR combines normalized turn-level disagreement $\tilde d_k$ and uncertainty $\tilde h_k$ by Soft-OR, $w_k = 1 - (1 - \tilde d_k)(1 - \tilde h_k)$, so that either signal can preserve a larger relative distillation weight for a turn.}
\label{fig:tdur-motivation}
\end{figure*}

Language agents extend large language models from static text generation to interactive decision making, where models must observe states, choose actions, call tools, and revise plans over multiple turns~\citep{yao2023react,schick2023toolformer}. Such capabilities make LLMs applicable to embodied instruction following, web shopping, search-augmented question answering, code execution, and other real-world tasks~\citep{shridhar2021alfworld,yao2022webshop,jin2025searchr1}. However, strong agentic behavior is usually concentrated in large models, whose inference cost and deployment overhead are undesirable in latency-, privacy-, and resource-sensitive settings. A practical goal is therefore to transfer multi-turn agentic competence from a stronger teacher to a smaller student, while still enabling the student to seek reward-driven improvements beyond pure teacher imitation.

Reinforcement learning (RL) has been widely used for post-training language agents. It optimizes the policy with environment-defined rewards, typically through PPO- or GRPO-style policy optimization~\citep{schulman2017ppo,shao2024deepseekmath,deepseek2025r1,yu2025dapo}, with recent agent-oriented variants adapting this idea to long-horizon interaction~\citep{zhou2024archer,qi2025webrl,feng2025gigpo,dong2025arpo}. More recently, on-policy distillation (OPD) has emerged as another promising paradigm: it trains the student on its own sampled trajectories while using a stronger teacher's token-level distribution as dense supervision~\citep{agarwal2024onpolicy,gu2024minillm,jin2026eopd,ye2026opcd,jang2026veto,yang2026gopd}.

However, RL and OPD exhibit complementary trade-offs in multi-turn agent training. RL directly pursues environment rewards and can improve beyond imitation, but sparse delayed feedback makes early exploration inefficient for small students. OPD provides dense token-level guidance and bootstraps learning quickly, but strong imitation can plateau near the teacher and suppress reward-improving deviations. This motivates an \textbf{annealed OPD--RL schedule}: OPD dominates early training to approach teacher-level behavior, while RL is gradually strengthened to drive reward-based exploration. The validation dynamics in \cref{fig:annealing-intuition} illustrate this complementarity: OPD improves quickly but saturates, GRPO grows slowly under sparse rewards, and a hard switch from OPD to GRPO after 80 steps remains below \method{}. This suggests that early RL signals can refine OPD guidance even during bootstrapping, while \textbf{soft annealing} avoids an abrupt objective shift and yields more stable improvement toward a higher ceiling.

A second challenge is deciding how to allocate dense teacher supervision within a long agent trajectory. Although OPD provides fine-grained token-level guidance, tokens are not the natural unit for allocating supervision across an interaction: an agent commits to a reasoning step, tool call, or environment action through a complete response turn, and environment feedback arrives between turns. Thus, token-level signals are well suited to delivering corrections within a response, whereas turns are the natural units for deciding which agent decisions should receive greater relative emphasis. A direct approach is to aggregate teacher--student disagreement within each turn and assign larger relative weights to turns with larger gaps. As shown in \cref{fig:tdur-motivation}(a), shallow turns consistently exhibit larger gaps than deep turns. This bias is useful early in training, when weak students often make their first mistakes near the beginning of a rollout; correcting these early departures can prevent errors from propagating into later states and contaminating the remaining context~\citep{wang2026tcod,zhong2026sod}.

However, observed disagreement is not a complete proxy for learning need throughout training. The key issue is that these gaps are measured on student rollouts: at deep turns, both models are conditioned on a student-generated prefix that may already contain accumulated errors. Such a drifted prefix can compress the observed gap even when the student has not mastered the underlying decision. Therefore, a small deep-turn gap does not necessarily indicate successful learning. Indeed, \cref{fig:tdur-motivation}(b) shows that as OPD progresses, shallow turns become more confident while deeper turns remain uncertain. We therefore complement disagreement with student uncertainty. Consistent with the token-level findings of TIP~\citep{xu2026tip}, the two signals capture distinct failure modes: disagreement identifies turns where the student departs from the teacher, including confident deviations, while uncertainty recovers fragile turns whose observed gap is small. \tdur{} aggregates sampled-token log-probabilities into \textbf{turn-level disagreement and uncertainty}, normalizes both within each trajectory, and fuses them with \textbf{Soft-OR}. It preserves more of the token-level OPD signal when either score is high and strongly attenuates a turn only when both scores are low, thereby extending coverage beyond disagreement-only weighting without discarding confident teacher--student deviations.

\begin{figure*}[t]
\centering
\includegraphics[width=\textwidth]{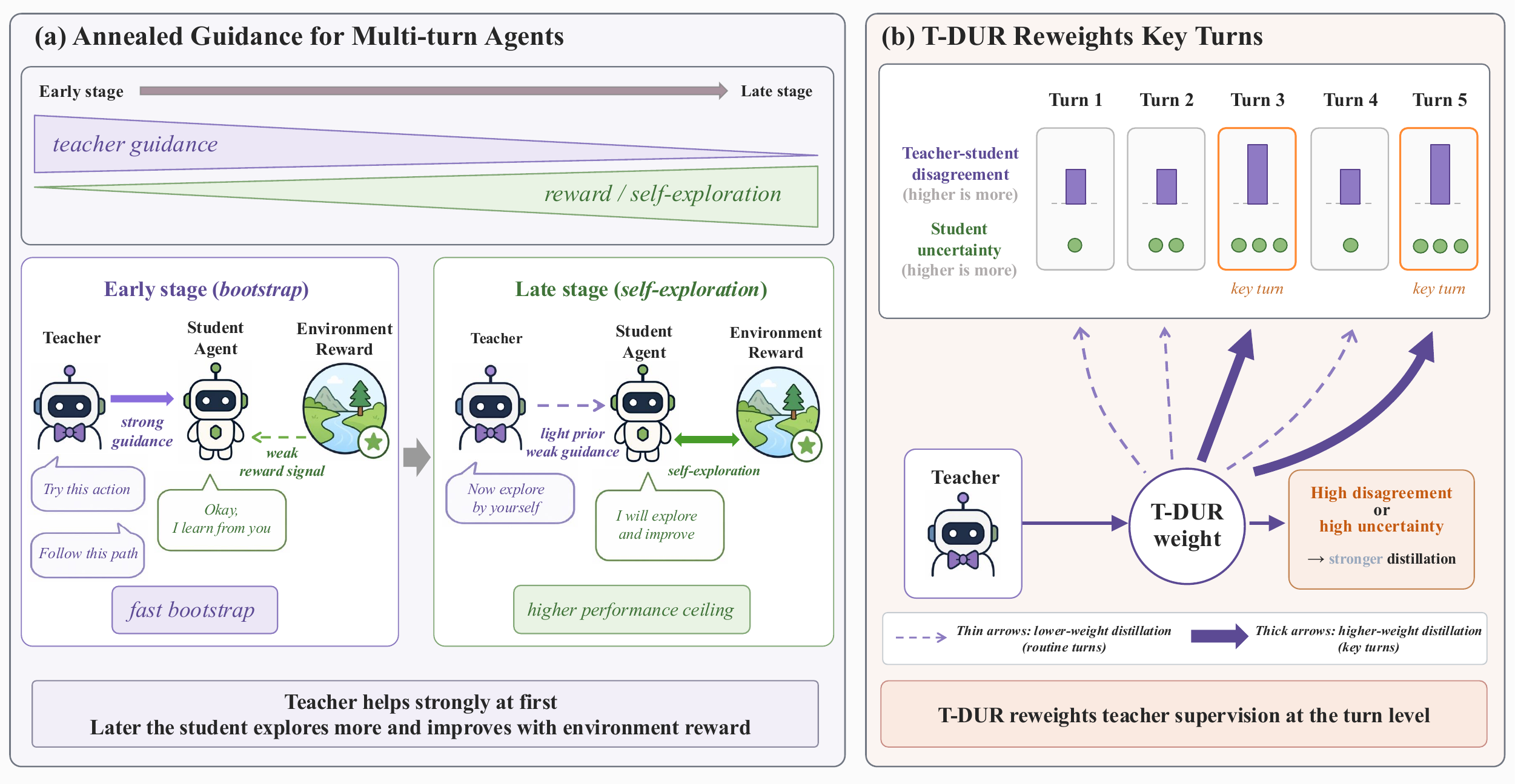}
\caption{Overview of \method{}. (a) Annealed guidance shifts from teacher-guided bootstrapping to reward-driven exploration. (b) T-DUR uses disagreement and uncertainty to selectively attenuate low-priority turns.}
\label{fig:atod-framework}
\end{figure*}

We propose \textbf{\method{}} (\textbf{A}nnealed \textbf{T}urn-aware \textbf{O}n-policy \textbf{D}istillation), a hybrid online distillation algorithm that addresses these two challenges with an \textbf{annealed OPD--RL schedule} and \textbf{Turn-level Disagreement-Uncertainty Reweighting (\tdur{})}, as illustrated in \cref{fig:atod-framework}. Together, these components improve early convergence, stabilize dense supervision, and raise the reward-driven performance ceiling. Our contributions are three-fold:
\begin{itemize}[leftmargin=*]
  \item We introduce an annealed OPD--RL training objective for multi-turn agent training. By letting OPD dominate early and RL dominate later, \method{} addresses the exploration inefficiency of sparse-reward RL while preserving reward-driven improvement beyond pure teacher imitation.
  \item We propose Turn-level Disagreement-Uncertainty Reweighting (\tdur{}), a soft distillation gating mechanism for agentic OPD. \tdur{} uses student uncertainty and teacher--student disagreement to selectively attenuate supervision on routine turns while preserving larger relative weights on turns that score high on either signal.
  \item We validate \method{} on ALFWorld, WebShop, and Search-QA across multiple student sizes. \method{} achieves the highest average success rate at every student scale, improving over both OPD and GRPO, and can surpass the corresponding teacher models on ALFWorld and WebShop.
\end{itemize}

\section{Related Work}
\label{sec:related-work}

\paragraph{Language agents and tool-integrated reasoning.}
Language agents combine reasoning with action, enabling LLMs to interact with tools, environments, and external feedback~\citep{yao2023react,schick2023toolformer}. Benchmarks such as ALFWorld, WebShop, and search-augmented QA require multi-step decision making under partial observations and delayed rewards~\citep{shridhar2021alfworld,yao2022webshop,jin2025searchr1}. These settings differ from single-turn text generation because errors can propagate across turns and because many actions are routine while a few are decisive.

\paragraph{Reinforcement learning for language models.}
Policy optimization methods, including PPO and GRPO, have been widely used to improve language-model reasoning and alignment~\citep{schulman2017ppo,shao2024deepseekmath,deepseek2025r1,yu2025dapo}. Agent-specific RL methods further address long-horizon interaction through hierarchical multi-turn optimization, online curricula, group-in-group credit assignment, and adaptive exploration~\citep{zhou2024archer,qi2025webrl,feng2025gigpo,dong2025arpo}. GRPO is attractive because it estimates relative advantages within sampled groups without requiring a separate value model. However, in long-horizon agent tasks, sparse terminal rewards and high-variance exploration can make pure RL inefficient, especially for small models.

\paragraph{On-policy distillation.}
OPD trains the student on its own generated trajectories while using a teacher distribution as dense supervision~\citep{agarwal2024onpolicy,gu2024minillm}. This reduces the train--test mismatch of offline supervised distillation and improves sample efficiency. Recent OPD variants improve stability, context transfer, entropy handling, or teacher-ceiling behavior~\citep{jin2026eopd,jang2026veto,ye2026opcd,yang2026gopd}. For tool-using or multi-turn agents, recent work has explored step-wise, temporal, self-distilled, skill-conditioned, or importance-aware distillation to mitigate unreliable or redundant teacher signals~\citep{zhong2026sod,wang2026tcod,zhao2026opsd,wang2026skillsd,lu2026sdar,xu2026tip}. \method{} is complementary: it combines OPD with reward-driven GRPO through a dynamic schedule and uses T-DUR to softly gate only the distillation component.

\section{Preliminaries}
\label{sec:preliminaries}

\subsection{Multi-Turn Agent Trajectories}

We consider post-training a small language-model agent for multi-turn interactive tasks. Given an input $x$, the student policy $\student$ interacts with external feedback over $K$ steps. At step $k$, the model generates a response $a_k$, which may contain reasoning, an environment action or tool invocation, or a final answer. If an external action is taken, an observation $o_k$ is appended to the context and conditions subsequent generations.

A trajectory is defined as
\begin{equation}
\traj = (x, a_1, o_1, \ldots, a_K, o_K, a_{K+1}),
\label{eq:trajectory}
\end{equation}
where $a_{K+1}$ denotes the final response. The policy generates only model tokens, while observations are provided by the environment. Let $a_t$ denote a generated token and $s_t$ its prefix context, which may include both previous model outputs and environment observations.

\subsection{Group Relative Policy Optimization}

Group Relative Policy Optimization (GRPO) is a reinforcement learning algorithm that updates the policy using relative rewards within a group of sampled trajectories~\citep{shao2024deepseekmath}. We assume access to an outcome-level reward function $R(\traj)$ defined on complete trajectories. For each input $x$, a group of trajectories $\{\traj_i\}_{i=1}^{G}$ is sampled from the old policy $\oldstudent$, each receiving reward $r_i=R(\traj_i)$. The group-relative advantage is computed as
\begin{equation}
\hat A_i^{\grpo}
= \frac{r_i-\mathrm{mean}(\{r_j\}_{j=1}^{G})}{\mathrm{std}(\{r_j\}_{j=1}^{G})+\epsilon_A}.
\label{eq:grpo-advantage}
\end{equation}
Let
\begin{equation}
\eta_{i,t}(\theta)=\frac{\student^{\theta}(a_{i,t}\mid s_{i,t})}{\oldstudent(a_{i,t}\mid s_{i,t})}
\label{eq:ratio}
\end{equation}
denote the token-level importance ratio. The GRPO objective is
\begin{equation}
\resizebox{0.92\columnwidth}{!}{$
\begin{aligned}
\loss_{\grpo}(\theta)
  &= -\mathbb{E}_{x}\Bigl[\frac{1}{G}\sum_{i=1}^{G}\frac{1}{|\mathcal{T}_i|}\sum_{t\in\mathcal{T}_i}\min\Bigl(\eta_{i,t}(\theta)\hat A_i^{\grpo}, \\
  &\qquad \mathrm{clip}(\eta_{i,t}(\theta),1-\epsilon,1+\epsilon)\hat A_i^{\grpo}\Bigr)\Bigr],
\end{aligned}
$}
\label{eq:grpo-loss}
\end{equation}
where $\mathcal{T}_i$ denotes the model-generated token positions in trajectory $\traj_i$. Equivalently, each generated token in $\traj_i$ receives $A_t^{\grpo}=\hat A_i^{\grpo}$. This objective provides an on-policy learning signal based on relative trajectory performance, but it relies on sparse outcome-level rewards.

\subsection{On-Policy Distillation}

On-policy distillation (OPD) is a post-training paradigm that provides dense token-level supervision on student-generated trajectories by aligning the student policy with a frozen teacher distribution~\citep{agarwal2024onpolicy,gu2024minillm}. Given trajectories $\{\traj_i\}_{i=1}^{G}$ sampled from the old student policy, the OPD objective can be written as
\begin{multline}
\loss_{\opd}(\theta) = \mathbb{E}_{x}\Bigg[
  \frac{1}{G}\sum_{i=1}^{G}\frac{1}{|\mathcal{T}_i|}
  \sum_{t\in\mathcal{T}_i}
  \eta_{i,t}(\theta) \\
  \Bigl(\log\student^{\theta}(a_{i,t}\mid s_{i,t})
  -\log\teacher(a_{i,t}\mid s_{i,t})\Bigr)
  \Bigg],
\label{eq:opd-objective}
\end{multline}
where $\mathcal{T}_i$ denotes the model-generated token positions in trajectory $\traj_i$, and $\eta_{i,t}(\theta)$ is the same importance ratio as in \cref{eq:ratio}. This objective is a sampled estimator of the reverse KL divergence from the student policy to the teacher policy on student-visited states.

Equivalently, for a generated token $a_t$, OPD provides the token-level distillation signal
\begin{equation}
\Delta\log p_t = \log\teacher(a_t\mid s_t)-\log\student(a_t\mid s_t),
\label{eq:delta-logp}
\end{equation}
which is positive when the teacher assigns higher probability to the sampled token than the student. In advantage form, pure OPD uses
\begin{equation}
A^{\opd}_t = \Delta\log p_t.
\label{eq:opd-advantage}
\end{equation}

\section{Method}
\label{sec:method}

\subsection{Overview}

\method{} combines two complementary learning signals. The teacher provides dense token-level guidance through OPD, which is especially useful when the student is still weak. The environment reward provides the task-defining signal through GRPO, which is essential for correcting teacher bias and improving beyond imitation. Instead of choosing one signal, \method{} uses both in a single token-level advantage:
\begin{equation}
A_t = \kappa(s)\,A^{\opd}_t + \rho(s)\,A^{\grpo}_t,
\label{eq:hybrid-advantage}
\end{equation}
where
\begin{equation}
A^{\opd}_t = \Delta\log p_t\,w_{k(t)}.
\label{eq:expanded-opd-advantage}
\end{equation}
Here $s$ is the global training step, and $k(t)$ maps token position $t$ to the turn that contains it. The coefficient $\kappa(s)$ controls how much the update follows the teacher, while $\rho(s)$ controls how much it follows the environment reward. The scalar $w_{k(t)}$ is the \tdur{} weight of token $t$'s turn. It only reweights the OPD term, so it changes where teacher supervision is applied without weakening the reward signal. The actor is then optimized with the same clipped surrogate as GRPO in \cref{eq:grpo-loss}, using $A_t$ as the advantage. We do not add an explicit KL penalty; the teacher effect is already included through $A^{\opd}_t$.

\subsection{Dynamic OPD/RL Coefficient Annealing}
\label{sec:annealing}

A fixed OPD/RL mixture is suboptimal because the two signals are useful at different stages. Early in training, rewards are sparse and noisy, while teacher guidance quickly teaches the student valid actions and interaction patterns. Later, pure imitation becomes limiting: the student may inherit teacher mistakes and cannot easily exceed the teacher. Therefore, \method{} gradually shifts the update from teacher-guided learning to reward-guided learning.

We use a simple progress variable
\begin{equation}
p(s)=\min\Bigl(\frac{s}{T},1\Bigr),
\label{eq:progress}
\end{equation}
where $s$ is the current training step and $T$ is the total annealing steps, so $p(s)$ increases from $0$ to $1$ during the annealing window. Since $p(s)\in[0,1]$, both coefficients are simple linear interpolations: $\kappa(s)$ decreases from $\kappa_{\mathrm{init}}$ to $\kappa_{\min}$, while $\rho(s)$ increases from $\rho_{\mathrm{init}}$ to $\rho_{\max}$:
\begin{equation}
\begin{aligned}
\kappa(s)&=\kappa_{\mathrm{init}}-(\kappa_{\mathrm{init}}-\kappa_{\min})\,p(s), \\
\rho(s)&=\rho_{\mathrm{init}}+(\rho_{\max}-\rho_{\mathrm{init}})\,p(s).
\end{aligned}
\label{eq:kappa-rho}
\end{equation}
Intuitively, $\kappa(s)$ dominates early, letting teacher guidance bootstrap the student when reward feedback is still sparse; $\rho(s)$ dominates later, shifting the update toward reward-driven exploration. Setting $\kappa_{\min}>0$ keeps a weak teacher anchor that prevents severe drift, but does not prevent reward-driven improvement.

\begin{table*}[t]
\centering
\caption{Performance on ALFWorld, Search-QA, and WebShop. We report success rate (SR, \%) and average trajectory length (Len.). Avg. SR is the mean success rate over the three datasets. Vanilla denotes the untrained base model. Methods marked with $^\dagger$ do not use a teacher model.}
\label{tab:main-results}
\footnotesize
\setlength{\tabcolsep}{7.8pt}
\setlength{\aboverulesep}{0.25ex}
\setlength{\belowrulesep}{0.25ex}
\setlength{\cmidrulesep}{0.12ex}
\renewcommand{\arraystretch}{0.94}
\begin{tabular}{cccccccc}
\toprule
\multirow{2}{*}{\textbf{Method}} & \multicolumn{2}{c}{\textbf{ALFWorld}} & \multicolumn{2}{c}{\textbf{Search-QA}} & \multicolumn{2}{c}{\textbf{WebShop}} & \multirow{2}{*}{\textbf{Avg. SR}} \\[-0.35ex]
\cmidrule(lr){2-3}\cmidrule(lr){4-5}\cmidrule(lr){6-7}
& \textbf{SR} & \textbf{Len.} & \textbf{SR} & \textbf{Len.} & \textbf{SR} & \textbf{Len.} & \\[-0.25ex]
\midrule
\rowcolor{yellow!20}
Qwen3-4B GRPO (Teacher) & 76.56 & 18.95 & 48.19 & 2.51 & 82.03 & 6.21 & 68.93 \\
\midrule
\rowcolor{gray!12}
\multicolumn{8}{c}{\textit{Qwen3-0.6B Student} (teacher: Qwen3-4B GRPO)} \\
Vanilla$^\dagger$ & 0.78 & 50.00 & 14.50 & 1.57 & 4.69 & 9.59 & 6.66 \\
GRPO$^\dagger$ & 30.47 & 44.13 & 39.36 & 2.69 & 29.69 & 9.73 & 33.17 \\
SDAR$^\dagger$ & 28.12 & 45.27 & 39.60 & 2.43 & 8.59 & 10.27 & 25.44 \\
OPD & \underline{76.56} & 18.48 & 41.46 & 2.77 & \underline{85.16} & 6.23 & \underline{67.73} \\
SOD & \underline{76.56} & 22.90 & 40.19 & 2.81 & 80.47 & 5.95 & 65.74 \\
TCOD & 74.22 & 20.15 & \underline{42.04} & 2.67 & 84.38 & \textbf{5.94} & 66.88 \\
\rowcolor{blue!10}
\textbf{\method{}} & \textbf{82.81} & \textbf{15.84} & \textbf{42.33} & 2.77 & \textbf{86.72} & 7.48 & \textbf{70.62} \\
\midrule
\rowcolor{gray!12}
\multicolumn{8}{c}{\textit{Qwen3-1.7B Student} (teacher: Qwen3-4B GRPO)} \\
Vanilla$^\dagger$ & 10.16 & 47.91 & 31.15 & 2.09 & 5.47 & 8.43 & 15.59 \\
GRPO$^\dagger$ & 31.25 & 40.38 & 41.99 & \textbf{2.64} & 47.66 & 10.01 & 40.30 \\
SDAR$^\dagger$ & 39.84 & 37.56 & 42.24 & 2.67 & 43.75 & 8.90 & 41.94 \\
OPD & 72.66 & \textbf{16.41} & 44.73 & 2.66 & 77.34 & 6.15 & 64.91 \\
SOD & \underline{75.78} & 27.95 & 44.43 & 2.69 & 76.56 & \textbf{5.59} & 65.59 \\
TCOD & 71.88 & 18.88 & \underline{45.02} & 2.74 & \underline{81.25} & 6.10 & \underline{66.05} \\
\rowcolor{blue!10}
\textbf{\method{}} & \textbf{80.47} & 22.69 & \textbf{45.21} & \textbf{2.64} & \textbf{89.06} & 6.20 & \textbf{71.58} \\
\midrule
\rowcolor{yellow!20}
Qwen3-30B-A3B GRPO (Teacher) & 80.47 & 23.53 & 50.49 & 2.65 & 75.78 & 8.02 & 68.91 \\
\midrule
\rowcolor{gray!12}
\multicolumn{8}{c}{\textit{Qwen3-4B Student} (teacher: Qwen3-30B-A3B GRPO)} \\
Vanilla$^\dagger$ & 24.22 & 42.43 & 32.71 & 1.77 & 0.78 & 13.99 & 19.24 \\
GRPO$^\dagger$ & 76.56 & 18.95 & 48.19 & \textbf{2.51} & \textbf{82.03} & \textbf{6.21} & \underline{68.93} \\
SDAR$^\dagger$ & 77.34 & 20.85 & 47.80 & 2.59 & 67.97 & 7.91 & 64.37 \\
OPD & \underline{80.47} & 21.12 & \underline{48.93} & 2.69 & 75.00 & 8.53 & 68.13 \\
SOD & 79.69 & 19.02 & 48.58 & 2.61 & 76.56 & 6.68 & 68.28 \\
TCOD & 76.56 & 23.45 & 48.39 & 2.57 & 75.78 & 9.21 & 66.91 \\
\rowcolor{blue!10}
\textbf{\method{}} & \textbf{85.16} & \textbf{16.75} & \textbf{49.12} & 2.62 & \underline{78.91} & 8.29 & \textbf{71.06} \\
\bottomrule
\end{tabular}
\end{table*}

\subsection{Turn-level Disagreement-Uncertainty Reweighting}
\label{sec:tdur}

For each turn $k$, \tdur{} computes two sampled-token statistics over response tokens.

\paragraph{Disagreement proxy.}
\begin{equation}
 d_k = \frac{1}{N_k}\sum_{t=1}^{N_k}\left|\log\student(a^{(k)}_t\mid s^{(k)}_t)-\log\teacher(a^{(k)}_t\mid s^{(k)}_t)\right|.
\label{eq:divergence-proxy}
\end{equation}
A high $d_k$ indicates strong student--teacher disagreement on the turn.

\paragraph{Uncertainty proxy.}
\begin{equation}
 h_k = \frac{1}{N_k}\sum_{t=1}^{N_k}\left(-\log\student(a^{(k)}_t\mid s^{(k)}_t)\right).
\label{eq:entropy-proxy}
\end{equation}
This sampled-token negative log-probability estimates the student's uncertainty over the turn.

\paragraph{Per-trajectory normalization.}
For turns within the same trajectory $\traj$, we normalize
\begin{equation}
\begin{aligned}
\tilde d_k&=\frac{d_k-\min_{j\in\traj}d_j}{\max_{j\in\traj}d_j-\min_{j\in\traj}d_j},\\
\tilde h_k&=\frac{h_k-\min_{j\in\traj}h_j}{\max_{j\in\traj}h_j-\min_{j\in\traj}h_j}.
\end{aligned}
\label{eq:minmax}
\end{equation}
If the denominator is below $10^{-8}$, the normalized value is set to $0.5$. Per-trajectory normalization avoids mixing scales across tasks, trajectory lengths, and environment states.

\paragraph{Soft-OR fusion.}
The final \tdur{} turn weight is denoted by $w_k$:
\begin{equation}
w_k = 1-(1-\tilde d_k)(1-\tilde h_k),\qquad w_k\in[0,1].
\label{eq:softor}
\end{equation}
This continuous weight becomes large when either student--teacher disagreement or student uncertainty is high. Because $w_k\in[0,1]$, \tdur{} acts as a soft gate: relative to uniform OPD with unit weights, it selectively attenuates turns that score low on both signals while preserving a larger fraction of the original distillation signal on uncertain or high-disagreement turns. The detailed training procedure for \method{} is provided in \cref{alg:atod} in Appendix~\ref{app:algorithm}.

\begin{figure*}[!t]
\centering
\begin{minipage}[t]{0.585\textwidth}
  \centering
  \includegraphics[width=\linewidth]{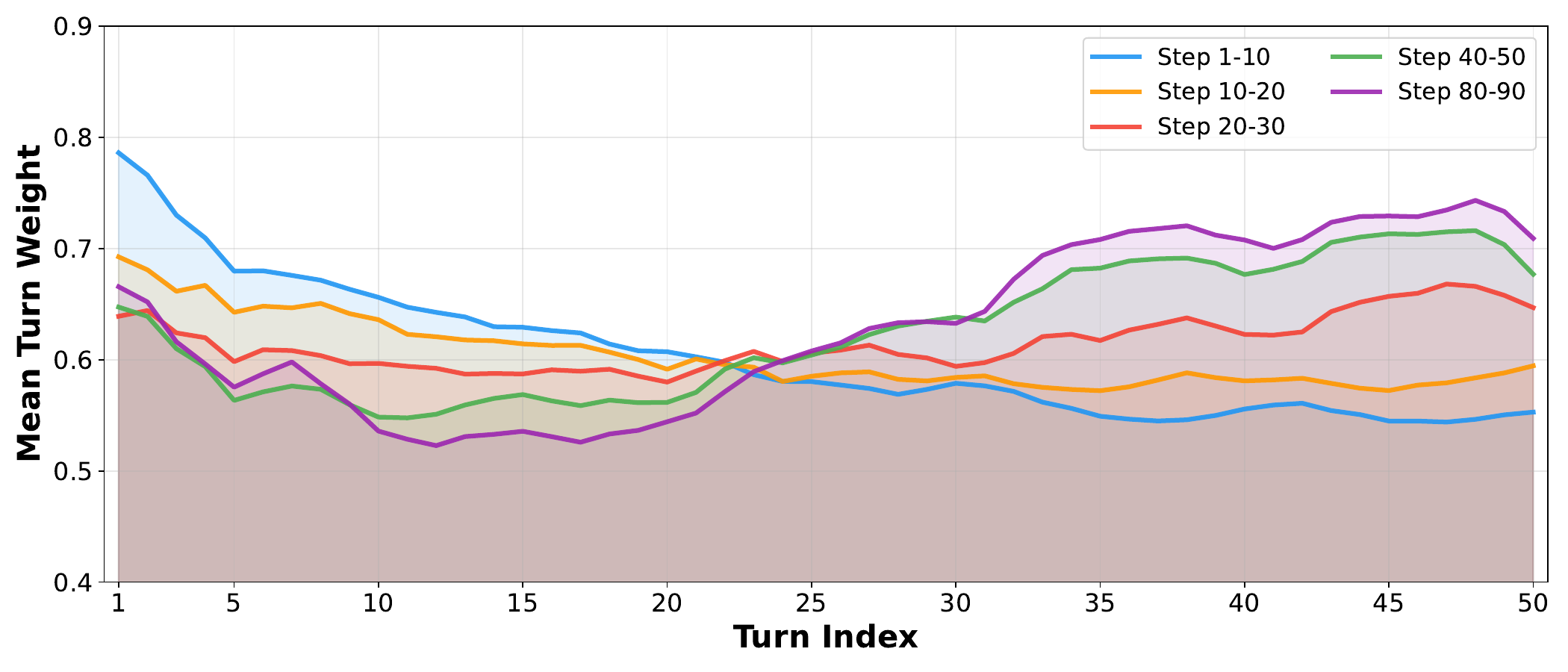}
  \par\smallskip\textbf{(a)} Training-stage comparison
\end{minipage}\hfill
\begin{minipage}[t]{0.395\textwidth}
  \centering
  \includegraphics[width=\linewidth]{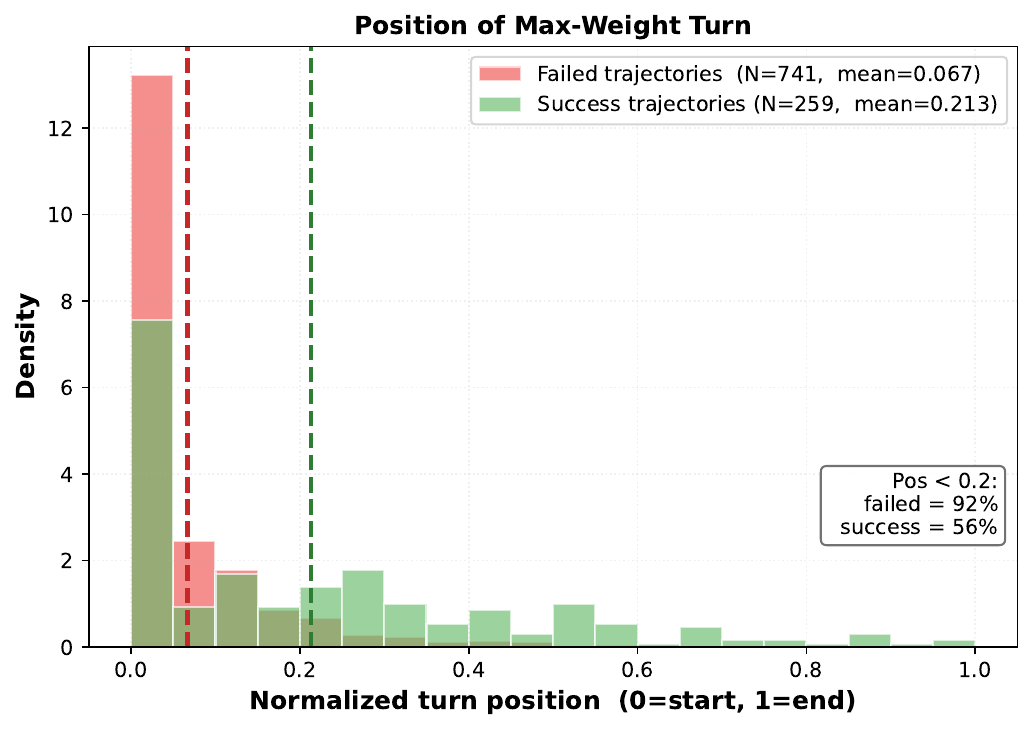}
  \par\smallskip\textbf{(b)} Outcome-conditioned critical-turn position
\end{minipage}
\caption{Turn-level \tdur{} diagnostics for the 1.7B student on ALFWorld. (a) Mean weights across training stages. (b) Normalized maximum-weight turn positions by outcome (0: start; 1: end); dashed lines mark means.}
\label{fig:tdur-weight-diagnostics}
\end{figure*}

\section{Experiments}
\label{sec:experiments}

\subsection{Experimental Setup}
\label{sec:experimental-setup}

\paragraph{Datasets \& Benchmarks.}
We evaluate \method{} on three long-horizon agent benchmarks: ALFWorld, WebShop, and Search-QA. ALFWorld tests embodied instruction following in text-based household environments, WebShop evaluates goal-conditioned web navigation and product selection, and Search-QA measures search-augmented question answering over open-domain and multi-hop tasks~\citep{shridhar2021alfworld,yao2022webshop,jin2025searchr1}. Following common Search-R1-style protocols, the Search-QA suite covers Natural Questions, TriviaQA, PopQA, HotpotQA, 2WikiMultiHopQA, MuSiQue, and Bamboogle~\citep{kwiatkowski2019natural,joshi2017triviaqa,yang2018hotpotqa,trivedi2022musique,press2023compositionality}. Together, these benchmarks cover embodied interaction, web interaction, and tool-assisted knowledge reasoning.

\paragraph{Evaluation Setups.}
We use Qwen3-0.6B, Qwen3-1.7B, and Qwen3-4B as student models. The 0.6B and 1.7B students use a Qwen3-4B model trained with GRPO as the teacher, while the 4B student uses a Qwen3-30B-A3B GRPO teacher (150-step checkpoint). Unless otherwise specified, each prompt is sampled on policy with group size $G{=}8$ so that GRPO-style group-relative advantages can be computed from environment feedback. We report validation success rate (\%) and the corresponding average trajectory length on ALFWorld, Search-QA, and WebShop. The main comparison uses a 150-step training window for all student sizes.

\begin{figure*}[!t]
\centering
\includegraphics[width=\textwidth]{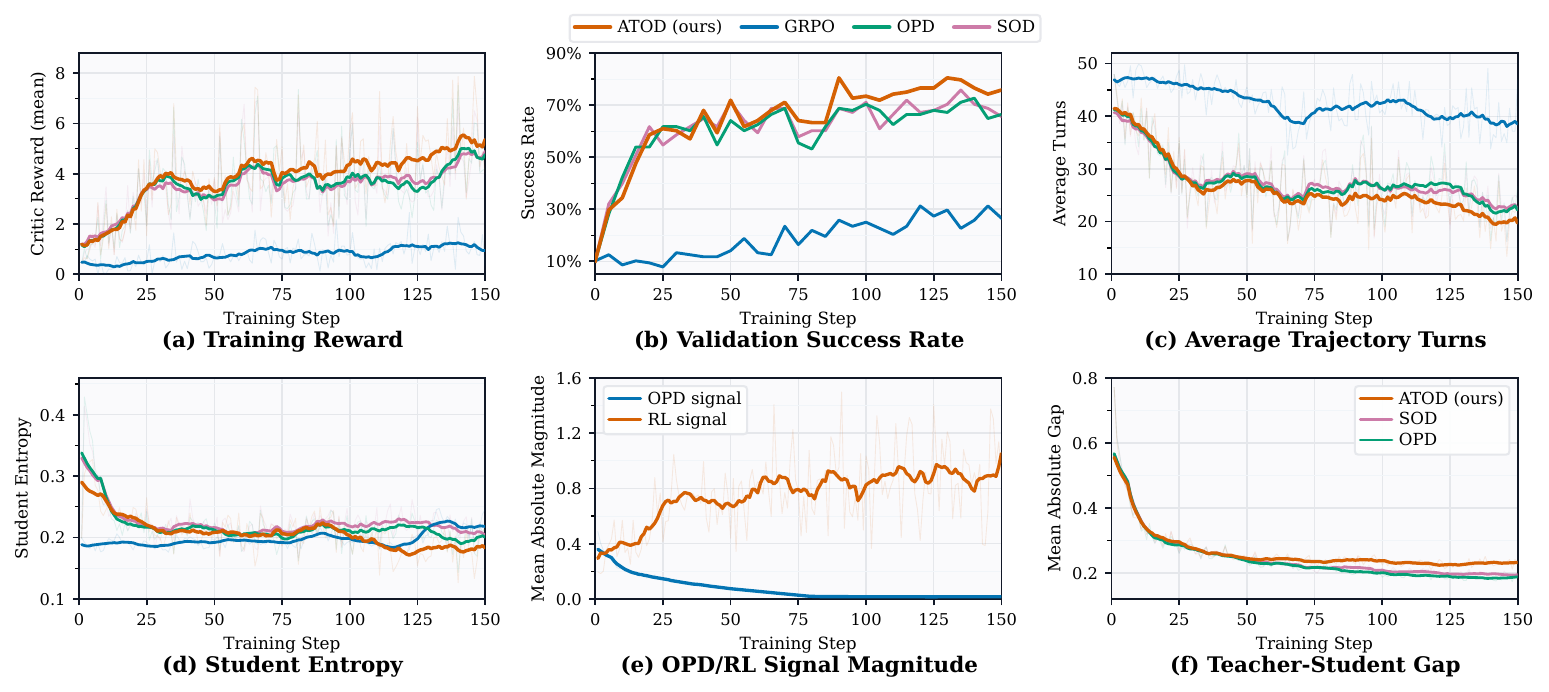}
\caption{Training dynamics and internal diagnostic metrics on ALFWorld. (a) Training reward. (b) Validation success rate. (c) Average trajectory turns. (d) Student entropy. (e) OPD/RL signal magnitude. (f) Teacher--student gap.}
\label{fig:training-curves}
\label{fig:internal-metrics}
\end{figure*}

\paragraph{Baselines.}
We compare \method{} against post-training baselines of different supervision types: the untrained Vanilla model, the RL method GRPO, the self-distillation + RL method SDAR~\citep{lu2026sdar}, standard on-policy distillation (OPD)~\citep{agarwal2024onpolicy,gu2024minillm}, and the agentic on-policy distillation methods SOD~\citep{zhong2026sod} and TCOD~\citep{wang2026tcod}. Baseline descriptions are provided in Appendix~\ref{app:baselines}.

\subsection{Main Results}

The performance comparison across ALFWorld, Search-QA, and WebShop is summarized in \cref{tab:main-results}, from which we draw the following key observations:

\begin{itemize}[leftmargin=*]
\item \textbf{Obs 1: \method{} achieves the best overall performance across benchmarks and student sizes, and even surpasses the corresponding teacher models.} Across all three student sizes (0.6B, 1.7B, and 4B), \method{} achieves the highest average success rate among all student methods, with margins of \textbf{2.89\%} (0.6B), \textbf{5.53\%} (1.7B), and \textbf{2.13\%} (4B) over the second-best baseline, and attains the best per-task success rate in eight of the nine benchmark--student combinations. Notably, \method{} surpasses its teacher on ALFWorld and WebShop for all student sizes; on Search-QA it remains the strongest student-side method, with the gap to the teacher shrinking from \textbf{5.9} points (0.6B) to \textbf{1.4} points (4B) as the student scale grows.

\item \textbf{Obs 2: \method{} delivers dramatic improvements for smaller models with weak initial performance.} For the 0.6B student, which starts with near-zero success on ALFWorld (0.78\%) and WebShop (4.69\%), \method{} lifts performance to \textbf{82.81\%} and \textbf{86.72\%} respectively---a relative improvement of over \textbf{100$\times$} on ALFWorld compared to the vanilla model. In contrast, GRPO achieves only 30.47\% on ALFWorld, highlighting \method{}'s superior efficiency in the early training stage when sparse rewards make pure RL ineffective.

\end{itemize}

\begin{table}[ht!]
\centering
\caption{Ablation results on ALFWorld and WebShop. We report success rate (\%). The model columns denote Qwen3-0.6B, Qwen3-1.7B, and Qwen3-4B, respectively. \textbf{Avg.} is the mean across all six dataset/model combinations.}
\label{tab:ablation-results}
\footnotesize
\setlength{\tabcolsep}{2.8pt}
\renewcommand{\arraystretch}{1.12}
\begin{tabular}{@{}L{0.22\columnwidth}ccc@{\hspace{0.02\columnwidth}}ccc@{\hspace{0.015\columnwidth}}c@{}}
\toprule
\multirow{2}{*}{\textbf{Ablation}} & \multicolumn{3}{c}{\textbf{ALFWorld SR (\%)}} & \multicolumn{3}{c}{\textbf{WebShop SR (\%)}} & \multirow{2}{*}{\textbf{Avg.}} \\
\cmidrule(lr){2-4}\cmidrule(lr){5-7}
& \textbf{0.6B} & \textbf{1.7B} & \textbf{4B} & \textbf{0.6B} & \textbf{1.7B} & \textbf{4B} \\
\midrule
Token-RW & 78.91 & 79.69 & 83.59 & 77.34 & 85.94 & 71.88 & 79.56 \\
Entropy-RW & 82.03 & 80.47 & 84.38 & 78.13 & 87.50 & 76.56 & 81.51 \\
Div-RW & 80.47 & 79.69 & 84.38 & 65.63 & 77.34 & 75.78 & 77.22 \\
w/o T-DUR & 78.91 & 75.78 & 84.38 & 85.94 & 85.94 & 73.44 & 80.73 \\
\mbox{w/o Annealing} & 75.78 & 73.44 & 83.59 & 49.22 & 87.50 & 75.00 & 74.09 \\
\midrule
\multicolumn{8}{@{}c@{}}{\begingroup\setlength{\fboxsep}{0pt}\colorbox{blue!10}{\begin{tabular}{@{}L{0.22\columnwidth}ccc@{\hspace{0.02\columnwidth}}ccc@{\hspace{0.015\columnwidth}}c@{}}\textbf{\method{}} & \textbf{82.81} & \textbf{80.47} & \textbf{85.16} & \textbf{86.72} & \textbf{89.06} & \textbf{78.91} & \textbf{83.86} \\[-0.15ex]\end{tabular}}\endgroup} \\
\bottomrule
\end{tabular}
\end{table}

\subsection{Ablation Study}
\label{sec:ablation-study}

To evaluate the contribution of each component in \method{}, we conduct five ablation studies on ALFWorld and WebShop across all student sizes (0.6B, 1.7B, and 4B): \textbf{(1)} applying the disagreement-uncertainty reweighting at the token level instead of the turn level (\textit{Token-RW}); \textbf{(2)} using entropy alone for reweighting (\textit{Entropy-RW}); \textbf{(3)} using disagreement alone for reweighting (\textit{Div-RW}); \textbf{(4)} removing the T-DUR reweighting entirely and using uniform turn weights (\textit{w/o T-DUR}); and \textbf{(5)} disabling the annealing schedule by fixing $\kappa(s)$ and $\rho(s)$ to constant values (\textit{w/o Annealing}). Here, ``RW'' denotes reweighting and ``Div'' denotes disagreement. The ALFWorld and WebShop results are shown in \cref{tab:ablation-results}. We additionally conduct an annealing-parameter ablation by varying the OPD/RL coefficient schedules; its settings and results are provided in Appendix~\ref{app:annealing-ablation}.

\begin{itemize}[leftmargin=*]
\item \textbf{Obs 3: The annealing schedule is crucial for combining imitation and reward optimization.} Removing the annealing mechanism leads to the most pronounced degradation across all student sizes, with the largest drop on relatively weaker students (e.g., 0.6B drops from 82.8\% to 75.8\%). This confirms that a fixed OPD/RL mixture fails to balance the two signals effectively: early training requires strong teacher guidance for bootstrapping, while later training needs reward-driven pressure to push beyond the teacher ceiling.

\item \textbf{Obs 4: Turn-level, dual-signal T-DUR is more robust than token-level, uniform, and single-signal reweighting.} Token-RW and uniform turn weighting reduce the average success rate by 4.30 and 3.13 points relative to \method{}, supporting complete response turns as a more semantically meaningful reweighting unit than individual tokens. The single-signal ablations show that both disagreement and uncertainty are necessary: \method{} improves over Entropy-RW and Div-RW by 2.35 and 6.64 points in overall average, and disagreement-only reweighting is particularly fragile on WebShop, dropping to 65.63\% on the 0.6B student. These results indicate that aggregating both signals over complete turns yields more consistent performance: token-level weights are susceptible to per-token fluctuations, uniform weights squander supervision on routine turns, and either signal alone misses useful decision steps.
\end{itemize}

\subsection{Training Dynamics and Diagnostic Metrics}
\label{sec:training-dynamics}

\cref{fig:tdur-weight-diagnostics}(a) shows that \tdur{} shifts relative emphasis from shallow turns early in training to difficult deep turns later. In \cref{fig:tdur-weight-diagnostics}(b), failed trajectories' highest-weight turns occur much earlier than successful trajectories': their mean normalized position is 0.067 vs.\ 0.213, and 92\% of failures have their highest-weight turn in the first 20\% of the episode, versus 56\% of successes. This indicates that early critical decisions often determine failure, and \tdur{} automatically concentrates weight on these decisive turns.

\cref{fig:training-curves} shows that \method{} combines the rapid early improvement of distillation-based methods with continued gains later in training, ultimately achieving the highest validation success rate. It also produces shorter trajectories than GRPO, indicating more efficient interaction. The OPD signal is strongest early and gradually decays, while the RL signal remains strong to support later reward optimization. During the first half of training, \method{}, OPD, and SOD reduce the teacher--student gap similarly; later, \method{} maintains a larger gap as annealing weakens imitation and preserves room for reward-driven behavior beyond the teacher.

\section{Conclusion}
\label{sec:conclusion}

We presented \method{}, an annealed turn-aware on-policy distillation algorithm for language agents. \method{} addresses the exploration inefficiency of pure RL, the teacher-ceiling problem of pure OPD, and the signal dilution caused by uniform turn weighting. By encoding OPD and GRPO into a single hybrid advantage, annealing their coefficients over training, and applying T-DUR only to teacher supervision, the method provides a simple and implementation-friendly path toward training compact agents that first imitate and then continuously self-improve via environmental feedback.

\bibliography{references}

@article{zhong2026sod,
  title   = {SOD: Step-wise On-policy Distillation for Small Language Model Agents},
  author  = {Zhong, Qiyong and Zheng, Mao and Song, Mingyang and Lin, Xin and Sun, Jie and Jiang, Houcheng and Wang, Xiang and Fang, Junfeng},
  journal = {arXiv preprint arXiv:2605.07725},
  year    = {2026}
}

@inproceedings{yao2023react,
  title     = {ReAct: Synergizing Reasoning and Acting in Language Models},
  author    = {Yao, Shunyu and Zhao, Jeffrey and Yu, Dian and Du, Nan and Shafran, Izhak and Narasimhan, Karthik R. and Cao, Yuan},
  booktitle = {International Conference on Learning Representations},
  year      = {2023}
}

@inproceedings{schick2023toolformer,
  title     = {Toolformer: Language Models Can Teach Themselves to Use Tools},
  author    = {Schick, Timo and Dwivedi-Yu, Jane and Dess\`i, Roberto and Raileanu, Roberta and Lomeli, Maria and Hambro, Eric and Zettlemoyer, Luke and Cancedda, Nicola and Scialom, Thomas},
  booktitle = {Advances in Neural Information Processing Systems},
  year      = {2023}
}

@inproceedings{agarwal2024onpolicy,
  title     = {On-Policy Distillation of Language Models: Learning from Self-Generated Mistakes},
  author    = {Agarwal, Rishabh and Vieillard, Nino and Zhou, Yongchao and Stanczyk, Piotr and Garea, Sabela Ramos and Geist, Matthieu and Bachem, Olivier},
  booktitle = {International Conference on Learning Representations},
  year      = {2024}
}

@inproceedings{gu2024minillm,
  title     = {MiniLLM: Knowledge Distillation of Large Language Models},
  author    = {Gu, Yuxian and Dong, Li and Wei, Furu and Huang, Minlie},
  booktitle = {International Conference on Learning Representations},
  year      = {2024}
}

@article{schulman2017ppo,
  title   = {Proximal Policy Optimization Algorithms},
  author  = {Schulman, John and Wolski, Filip and Dhariwal, Prafulla and Radford, Alec and Klimov, Oleg},
  journal = {arXiv preprint arXiv:1707.06347},
  year    = {2017}
}

@article{shao2024deepseekmath,
  title   = {DeepSeekMath: Pushing the Limits of Mathematical Reasoning in Open Language Models},
  author  = {Shao, Zhihong and Wang, Peiyi and Zhu, Qihao and Xu, Runxin and Song, Junxiao and others},
  journal = {arXiv preprint arXiv:2402.03300},
  year    = {2024}
}

@inproceedings{shridhar2021alfworld,
  title     = {ALFWorld: Aligning Text and Embodied Environments for Interactive Learning},
  author    = {Shridhar, Mohit and Yuan, Xingdi and C\^ot\'e, Marc-Alexandre and Bisk, Yonatan and Trischler, Adam and Hausknecht, Matthew},
  booktitle = {International Conference on Learning Representations},
  year      = {2021}
}

@inproceedings{yao2022webshop,
  title     = {WebShop: Towards Scalable Real-World Web Interaction with Grounded Language Agents},
  author    = {Yao, Shunyu and Chen, Howard and Yang, John and Narasimhan, Karthik},
  booktitle = {Advances in Neural Information Processing Systems},
  year      = {2022}
}

@article{wang2026tcod,
  title   = {TCOD: Exploring Temporal Curriculum in On-Policy Distillation for Multi-Turn Autonomous Agents},
  author  = {Wang, Jiaqi and Zhang, Wenhao and Shi, Weijie and Li, Yaliang and Cheng, James},
  journal = {arXiv preprint arXiv:2604.24005},
  year    = {2026}
}

@article{lu2026sdar,
  title   = {Self-Distilled Agentic Reinforcement Learning},
  author  = {Lu, Zhengxi and Yao, Zhiyuan and Han, Zhuowen and Wang, Zi-Han and Wu, Jinyang and Gu, Qi and Cai, Xunliang and Lu, Weiming and Xiao, Jun and Zhuang, Yueting and Shen, Yongliang},
  journal = {arXiv preprint arXiv:2605.15155},
  year    = {2026}
}

@article{xu2026tip,
  title   = {TIP: Token Importance in On-Policy Distillation},
  author  = {Xu, Yuanda and Sang, Hejian and Zhou, Zhengze and He, Ran and Wang, Zhipeng and Geramifard, Alborz},
  journal = {arXiv preprint arXiv:2604.14084},
  year    = {2026}
}

@article{jin2026eopd,
  title   = {Entropy-Aware On-Policy Distillation of Language Models},
  author  = {Jin, Woogyeol and Min, Taywon and Yang, Yongjin and Kadhe, Swanand Ravindra and Zhou, Yi and Wei, Dennis and Baracaldo, Nathalie and Lee, Kimin},
  journal = {arXiv preprint arXiv:2603.07079},
  year    = {2026}
}

@article{jang2026veto,
  title   = {Stable On-Policy Distillation through Adaptive Target Reformulation},
  author  = {Jang, Ijun and Yeom, Jewon and Yeo, Juan and Lim, Hyunggu and Kim, Taesup},
  journal = {arXiv preprint arXiv:2601.07155},
  year    = {2026}
}

@article{ye2026opcd,
  title   = {On-Policy Context Distillation for Language Models},
  author  = {Ye, Tianzhu and Dong, Li and Wu, Xun and Huang, Shaohan and Wei, Furu},
  journal = {arXiv preprint arXiv:2602.12275},
  year    = {2026}
}

@article{yang2026gopd,
  title   = {Learning beyond Teacher: Generalized On-Policy Distillation with Reward Extrapolation},
  author  = {Yang, Wenkai and Liu, Weijie and Xie, Ruobing and Yang, Kai and Yang, Saiyong and Lin, Yankai},
  journal = {arXiv preprint arXiv:2602.12125},
  year    = {2026}
}

@article{zhao2026opsd,
  title   = {Self-Distilled Reasoner: On-Policy Self-Distillation for Large Language Models},
  author  = {Zhao, Siyan and Xie, Zhihui and Liu, Mengchen and Huang, Jing and Pang, Guan and Chen, Feiyu and Grover, Aditya},
  journal = {arXiv preprint arXiv:2601.18734},
  year    = {2026}
}

@article{wang2026skillsd,
  title   = {Skill-SD: Skill-Conditioned Self-Distillation for Multi-Turn LLM Agents},
  author  = {Wang, Hao and Wang, Guozhi and Xiao, Han and Zhou, Yufeng and Pan, Yue and Wang, Jichao and Xu, Ke and Wen, Yafei and Ruan, Xiaohu and Chen, Xiaoxin and Qi, Honggang},
  journal = {arXiv preprint arXiv:2604.10674},
  year    = {2026}
}

@article{deepseek2025r1,
  title   = {DeepSeek-R1: Incentivizing Reasoning Capability in LLMs via Reinforcement Learning},
  author  = {{DeepSeek-AI} and Guo, Daya and Yang, Dejian and Zhang, Haowei and Song, Junxiao and Wang, Peiyi and Zhu, Qihao and Xu, Runxin and others},
  journal = {Nature},
  volume  = {645},
  pages   = {633--638},
  year    = {2025}
}

@article{yu2025dapo,
  title   = {DAPO: An Open-Source LLM Reinforcement Learning System at Scale},
  author  = {Yu, Qiying and Zhang, Zheng and Zhu, Ruofei and Yuan, Yufeng and Zuo, Xiaochen and Yue, Yu and Dai, Weinan and Fan, Tiantian and Liu, Gaohong and Liu, Lingjun and Liu, Xin and Lin, Haibin and Lin, Zhiqi and Ma, Bole and Sheng, Guangming and Tong, Yuxuan and Zhang, Chi and Zhang, Mofan and Zhang, Wang and Zhu, Hang and Zhu, Jinhua and Chen, Jiaze and Chen, Jiangjie and Wang, Chengyi and Yu, Hongli and Song, Yuxuan and Wei, Xiangpeng and Zhou, Hao and Liu, Jingjing and Ma, Wei-Ying and Zhang, Ya-Qin and Yan, Lin and Qiao, Mu and Wu, Yonghui and Wang, Mingxuan},
  journal = {arXiv preprint arXiv:2503.14476},
  year    = {2025}
}

@article{dong2025arpo,
  title   = {Agentic Reinforced Policy Optimization},
  author  = {Dong, Guanting and Mao, Hangyu and Ma, Kai and Bao, Licheng and Chen, Yifei and Wang, Zhongyuan and Chen, Zhongxia and Du, Jiazhen and Wang, Huiyang and Zhang, Fuzheng and Zhou, Guorui and Zhu, Yutao and Wen, Ji-Rong and Dou, Zhicheng},
  journal = {arXiv preprint arXiv:2507.19849},
  year    = {2025}
}

@article{feng2025gigpo,
  title   = {Group-in-Group Policy Optimization for LLM Agent Training},
  author  = {Feng, Lang and Xue, Zhenghai and Liu, Tingcong and An, Bo},
  journal = {arXiv preprint arXiv:2505.10978},
  year    = {2025}
}

@inproceedings{zhou2024archer,
  title     = {ArCHer: Training Language Model Agents via Hierarchical Multi-Turn RL},
  author    = {Zhou, Yifei and Zanette, Andrea and Pan, Jiayi and Levine, Sergey and Kumar, Aviral},
  booktitle = {Proceedings of the 41st International Conference on Machine Learning},
  pages     = {62178--62209},
  year      = {2024}
}

@inproceedings{qi2025webrl,
  title     = {WebRL: Training LLM Web Agents via Self-Evolving Online Curriculum Reinforcement Learning},
  author    = {Qi, Zehan and Liu, Xiao and Iong, Iat Long and Lai, Hanyu and Sun, Xueqiao and Zhao, Wenyi and Yang, Yu and Yang, Xinyue and Sun, Jiadai and Yao, Shuntian and Zhang, Tianjie and Xu, Wei and Tang, Jie and Dong, Yuxiao},
  booktitle = {International Conference on Learning Representations},
  year      = {2025}
}

@article{jin2025searchr1,
  title   = {Search-R1: Training LLMs to Reason and Leverage Search Engines with Reinforcement Learning},
  author  = {Jin, Bowen and Zeng, Hansi and Yue, Zhenrui and Yoon, Jinsung and Arik, Sercan and Wang, Dong and Zamani, Hamed and Han, Jiawei},
  journal = {arXiv preprint arXiv:2503.09516},
  year    = {2025}
}

@article{kwiatkowski2019natural,
  title   = {Natural Questions: A Benchmark for Question Answering Research},
  author  = {Kwiatkowski, Tom and Palomaki, Jennimaria and Redfield, Olivia and Collins, Michael and Parikh, Ankur and Alberti, Chris and Epstein, Danielle and Polosukhin, Illia and Devlin, Jacob and Lee, Kenton and Toutanova, Kristina and Jones, Llion and Kelcey, Matthew and Chang, Ming-Wei and Dai, Andrew M. and Uszkoreit, Jakob and Le, Quoc and Petrov, Slav},
  journal = {Transactions of the Association for Computational Linguistics},
  volume  = {7},
  pages   = {452--466},
  year    = {2019}
}

@inproceedings{joshi2017triviaqa,
  title     = {TriviaQA: A Large Scale Distantly Supervised Challenge Dataset for Reading Comprehension},
  author    = {Joshi, Mandar and Choi, Eunsol and Weld, Daniel and Zettlemoyer, Luke},
  booktitle = {Proceedings of the 55th Annual Meeting of the Association for Computational Linguistics},
  pages     = {1601--1611},
  year      = {2017}
}

@inproceedings{yang2018hotpotqa,
  title     = {HotpotQA: A Dataset for Diverse, Explainable Multi-Hop Question Answering},
  author    = {Yang, Zhilin and Qi, Peng and Zhang, Saizheng and Bengio, Yoshua and Cohen, William W. and Salakhutdinov, Ruslan and Manning, Christopher D.},
  booktitle = {Proceedings of the 2018 Conference on Empirical Methods in Natural Language Processing},
  pages     = {2369--2380},
  year      = {2018}
}

@article{trivedi2022musique,
  title   = {MuSiQue: Multihop Questions via Single-Hop Question Composition},
  author  = {Trivedi, Harsh and Balasubramanian, Niranjan and Khot, Tushar and Sabharwal, Ashish},
  journal = {Transactions of the Association for Computational Linguistics},
  volume  = {10},
  pages   = {539--554},
  year    = {2022}
}

@inproceedings{press2023compositionality,
  title     = {Measuring and Narrowing the Compositionality Gap in Language Models},
  author    = {Press, Ofir and Zhang, Muru and Min, Sewon and Schmidt, Ludwig and Smith, Noah A. and Lewis, Mike},
  booktitle = {Findings of the Association for Computational Linguistics: EMNLP 2023},
  year      = {2023}
}

\clearpage
\onecolumn
\raggedbottom
\appendix
\begin{center}
{\LARGE\bfseries Appendix}
\end{center}
\vspace{10pt}

\section*{Contents}
\begingroup
\setlength{\parindent}{0pt}
\newcommand{\AppendixMainEntry}[2]{%
  \noindent\textbf{\ref{#2}\quad #1}%
  \nobreak\hfill\nobreak
  \textbf{\pageref{#2}}\par\vspace{0.75em}%
}
\newcommand{\AppendixSubEntry}[2]{%
  \noindent\hspace{2.1em}\ref{#2}\quad #1%
  \nobreak\leaders\hbox{\normalfont\hbox to 0.75em{\hss.\hss}}\hfill\nobreak
  \pageref{#2}\par\vspace{0.35em}%
}
\AppendixMainEntry{Limitations}{app:limitations}
\AppendixMainEntry{Method and Algorithmic Details}{app:method-details}
\AppendixSubEntry{Complete Training Algorithm}{app:algorithm}
\AppendixSubEntry{Gradient Interpretation of the OPD Component}{app:gradient-interpretation}
\AppendixSubEntry{Conditional Targets and Concentration of T-DUR Proxies}{app:tdur-proxies}
\AppendixSubEntry{Properties of Soft-OR and Annealing}{app:softor-annealing}
\AppendixMainEntry{Experimental Setup and Supplementary Analyses}{app:experimental-details}
\AppendixSubEntry{Benchmarks}{app:benchmarks}
\AppendixSubEntry{Baselines}{app:baselines}
\AppendixSubEntry{Training Hyperparameters}{app:config}
\AppendixSubEntry{Annealing Parameter Ablation}{app:annealing-ablation}
\AppendixSubEntry{Additional Training Diagnostics}{app:additional-diagnostics}
\AppendixSubEntry{T-DUR Diagnostic Metrics on ALFWorld}{app:tdur-metrics}
\AppendixMainEntry{Qualitative Examples and Diagnostic Analyses}{app:qualitative-diagnostics}
\AppendixSubEntry{Environment Prompt Templates}{app:prompt-templates}
\AppendixSubEntry{Turn-level Case Studies}{app:case-studies}
\endgroup
\clearpage

\section{Limitations}
\label{app:limitations}

\paragraph{Evaluation scope.} We evaluate \method{} on text-based interactive benchmarks (ALFWorld, WebShop, Search-QA) with the Qwen3 model family as students. These benchmarks provide clean, turn-structured environments with well-defined rewards that isolate agentic decision dynamics without confounds from perception or open-ended tool interfaces, and Qwen3 offers strong, stable performance across scales, open availability, and wide adoption in recent OPD studies that facilitates fair comparison. Other agentic settings, such as GUI-based or multimodal web browsing, and other model families may exhibit different disagreement and uncertainty patterns and are worth exploring.

\paragraph{Dependence on teacher quality.} Like other teacher-based distillation methods, \method{}'s gains rely on a reasonably capable teacher: \tdur{} amplifies supervision precisely on turns where the teacher and student disagree or the student is uncertain, so the resulting benefit is harder to guarantee when the teacher itself is unreliable on such turns. Coupling \method{} with reliability-aware teacher gating is a promising direction for reducing this dependence. Due to resource constraints, we leave these explorations to future work.

\section{Method and Algorithmic Details}
\label{app:method-details}

This section collects the implementation-level and analytical details of \method{} that complement the main method section.

\subsection{Complete Training Algorithm}
\label{app:algorithm}

The complete \method{} training procedure is detailed below.

\begin{algorithm}[htbp]
\caption{\method{}: Annealed Turn-aware On-policy Distillation}
\label{alg:atod}
\begin{algorithmic}[1]
\Statex \textbf{Input:} Student policy $\student$, teacher policy $\teacher$, environment $\mathcal{E}$, prompt set $\mathcal{X}$, group size $G$, annealing schedule $(\kappa,\rho)$, clipping range $\epsilon$.
\Statex \textbf{Output:} Trained student policy $\student$.
\Statex
\Statex \textbf{Stage I: On-policy rollout and reward advantage}
\For{each prompt $x\in\mathcal{X}$}
  \State Sample $G$ trajectories $\{\traj_i\}_{i=1}^G$ with the old student policy $\oldstudent$ in $\mathcal{E}$.
  \State Evaluate each trajectory with the task reward $R(\traj_i)$.
  \State Compute group-relative GRPO advantages $A^{\grpo}$ using \cref{eq:grpo-advantage}.
\EndFor
\Statex
\Statex \textbf{Stage II: Turn-aware distillation reweighting}
\For{each trajectory $\traj$}
  \State Partition model-generated tokens into turns $\{\stepset_k\}_{k=1}^{K}$.
  \State Compute turn-level disagreement $d_k$ and uncertainty $h_k$ using \cref{eq:divergence-proxy,eq:entropy-proxy}.
  \State Fuse the normalized signals into T-DUR weights $\{w_k\}$ using \cref{eq:minmax,eq:softor}.
  \State Form the turn-weighted OPD advantage $A^{\opd}_t=\Delta\log p_t w_{k(t)}$ using \cref{eq:expanded-opd-advantage}.
\EndFor
\Statex
\Statex \textbf{Stage III: Annealed hybrid optimization}
\State Compute $\kappa(s)$ and $\rho(s)$ using \cref{eq:kappa-rho}.
\State Form the hybrid advantage:
\Statex \hspace{1.5em}$A_t=\kappa(s)A^{\opd}_t+\rho(s)A^{\grpo}_t$.
\State Update $\student$ by minimizing the clipped surrogate:
\Statex \hspace{1.5em}$\loss_{\mathrm{actor}}(\theta)=-\mathbb{E}_t\!\left[\min\!\left(\eta_t(\theta)A_t,\mathrm{clip}(\eta_t(\theta),1-\epsilon,1+\epsilon)A_t\right)\right]$.
\State Synchronize the old policy: $\oldstudent\leftarrow\student$.
\end{algorithmic}
\end{algorithm}

\subsection{Gradient Interpretation of the OPD Component}
\label{app:gradient-interpretation}

Inside the unclipped region where $\eta_t(\theta)\approx 1$, the distillation part of the actor loss is approximately
\begin{equation}
\loss_{\opd}(\theta)\approx -\mathbb{E}_t\left[\eta_t(\theta)\kappa(s)\Delta\log p_t w_{k(t)}\right].
\label{eq:opd-loss}
\end{equation}
Treating $\Delta\log p_t$ and the turn weight $w_{k(t)}$ as detached advantages, the gradient is
\begin{equation}
\nabla_\theta\loss_{\opd}\approx -\mathbb{E}_t\left[\kappa(s)\Delta\log p_t w_{k(t)}\nabla_\theta\log\student^\theta(a_t\mid s_t)\right].
\label{eq:opd-grad}
\end{equation}
Thus \method{} performs advantage-weighted likelihood ascent on student-sampled tokens, where the OPD advantage is high when the teacher assigns higher probability than the old student and T-DUR assigns a large disagreement-uncertainty weight to the turn.

\subsection{Conditional Targets and Concentration of T-DUR Proxies}
\label{app:tdur-proxies}

For a fixed state $s$, write $p_s=\student(\cdot\mid s)$ and $q_s=\teacher(\cdot\mid s)$, and define the absolute log-ratio discrepancy
\begin{equation}
D_{\mathrm{abs}}(p_s\Vert q_s)
:=\mathbb{E}_{a\sim p_s}\!\left[\left|\log\frac{p_s(a)}{q_s(a)}\right|\right].
\label{eq:absolute-log-ratio}
\end{equation}
This quantity is distinct from the reverse KL: it measures the expected magnitude of the sampled OPD correction without cancellation between positive and negative log-ratios.

\begin{proposition}
Consider a turn $k$ with a fixed token count $N_k$. Let $\mathcal{F}_{t-1}$ denote the history before sampling token $a_t$, so that $s_t$ is $\mathcal{F}_{t-1}$-measurable, and define
\begin{equation}
\bar H_k=\frac{1}{N_k}\sum_{t=1}^{N_k}H(p_{s_t}),
\qquad
\bar D^{\mathrm{abs}}_k=\frac{1}{N_k}\sum_{t=1}^{N_k}D_{\mathrm{abs}}(p_{s_t}\Vert q_{s_t}).
\end{equation}
Assume $q_{s_t}(a)>0$ whenever $p_{s_t}(a)>0$, and that the conditional variances of $-\log p_{s_t}(a_t)$ and $|\log(p_{s_t}(a_t)/q_{s_t}(a_t))|$ are bounded by $\sigma_h^2$ and $\sigma_d^2$, respectively. Then the T-DUR proxies satisfy
\begin{equation}
\mathbb{E}[h_k-\bar H_k]=0,
\qquad
\mathbb{E}[d_k-\bar D^{\mathrm{abs}}_k]=0,
\label{eq:proxy-unbiasedness}
\end{equation}
and
\begin{equation}
\mathbb{E}\!\left[(h_k-\bar H_k)^2\right]\leq\frac{\sigma_h^2}{N_k},
\qquad
\mathbb{E}\!\left[(d_k-\bar D^{\mathrm{abs}}_k)^2\right]\leq\frac{\sigma_d^2}{N_k}.
\label{eq:proxy-mse}
\end{equation}
Thus both turn-level proxies have estimation error of order $O(N_k^{-1/2})$, or mean-squared error of order $O(N_k^{-1})$, around their respective conditional targets.
\end{proposition}

\paragraph{Proof.}
Let
\begin{equation}
X_t=-\log p_{s_t}(a_t),
\qquad
Y_t=\left|\log\frac{p_{s_t}(a_t)}{q_{s_t}(a_t)}\right|.
\end{equation}
Because $a_t\sim p_{s_t}$, their conditional expectations are
\begin{equation}
\mathbb{E}[X_t\mid\mathcal{F}_{t-1}]=H(p_{s_t}),
\qquad
\mathbb{E}[Y_t\mid\mathcal{F}_{t-1}]=D_{\mathrm{abs}}(p_{s_t}\Vert q_{s_t}).
\end{equation}
Consequently, $X_t-H(p_{s_t})$ and $Y_t-D_{\mathrm{abs}}(p_{s_t}\Vert q_{s_t})$ are martingale differences. Martingale differences at distinct token positions are orthogonal, so averaging the conditional second-moment bounds over $N_k$ positions gives \cref{eq:proxy-unbiasedness,eq:proxy-mse}.

\paragraph{Relation to reverse KL.}
The signed log-ratio, rather than its absolute value, is the unbiased single-sample estimator of reverse KL:
\begin{equation}
\mathbb{E}_{a\sim p_s}\!\left[\log\frac{p_s(a)}{q_s(a)}\right]
=D_{\mathrm{KL}}(p_s\Vert q_s).
\label{eq:signed-kl-estimator}
\end{equation}
Accordingly, $d_k$ is not an unbiased estimator of reverse KL. It is unbiased for the average absolute log-ratio discrepancy in \cref{eq:absolute-log-ratio}, which obeys
\begin{equation}
D_{\mathrm{abs}}(p_s\Vert q_s)-D_{\mathrm{KL}}(p_s\Vert q_s)
=2\,\mathbb{E}_{a\sim p_s}\!\left[\left(\log\frac{q_s(a)}{p_s(a)}\right)_{+}\right]\geq0.
\label{eq:absolute-kl-relation}
\end{equation}
Since $|\Delta\log p_t|=|\log(p_{s_t}(a_t)/q_{s_t}(a_t))|$, $d_k$ has the direct interpretation of estimating the mean absolute magnitude of the OPD correction on turn $k$. The absolute value deliberately avoids cancellation and yields a non-negative signal for T-DUR gating; it is not used as an approximation claimed to be unbiased for the OPD reverse-KL objective.

\subsection{Properties of Soft-OR and Annealing}
\label{app:softor-annealing}

The function $f(d,h)=1-(1-d)(1-h)$ is monotone in both arguments, symmetric, satisfies $f(d,0)=d$ and $f(0,h)=h$, and is at least as large as each individual signal. It is the standard probabilistic OR t-conorm, making it appropriate when either uncertainty or disagreement should mark a turn as important.

The annealing functions in \cref{eq:kappa-rho} are Lipschitz in the training step with constants $|\kappa_{\mathrm{init}}-\kappa_{\min}|/T$ and $|\rho_{\max}-\rho_{\mathrm{init}}|/T$. This avoids abrupt changes in advantage scale and makes the transition between imitation and reward optimization smooth.

\section{Experimental Setup and Supplementary Analyses}
\label{app:experimental-details}

This section collects the experimental details needed to reproduce and interpret the reported results, together with supplementary ablation and training analyses.

\subsection{Benchmarks}
\label{app:benchmarks}

We evaluate \method{} on three text-based agent benchmarks that cover complementary interaction patterns and skill demands.

\paragraph{ALFWorld~\citep{shridhar2021alfworld}.} A text-based embodied household simulator, where the agent must accomplish multi-stage goals (e.g., finding, cleaning, and placing an object) by issuing discrete actions and reading textual observations. Tasks involve long trajectories with many routine navigation steps and a few decisive planning choices, making ALFWorld the primary testbed for long-horizon behavior in our study.

\paragraph{WebShop~\citep{yao2022webshop}.} A simulated online shopping environment, where the agent must locate and purchase a product that satisfies a detailed natural-language specification (e.g., attributes, size, color, and price) through searching, clicking, and option selection. It requires goal-conditioned web navigation and constraint checking over medium-length trajectories.

\paragraph{Search-QA~\citep{jin2025searchr1}.} A search-augmented question answering suite following the Search-R1 protocol, covering both open-domain and multi-hop questions from Natural Questions, TriviaQA, PopQA, HotpotQA, 2WikiMultiHopQA, MuSiQue, and Bamboogle~\citep{kwiatkowski2019natural,joshi2017triviaqa,yang2018hotpotqa,trivedi2022musique,press2023compositionality}. The agent alternates between issuing search queries and reasoning over retrieved results before producing a final answer, with relatively short trajectories centered on tool-assisted knowledge retrieval.

Together, these benchmarks span embodied interaction, web interaction, and tool-assisted knowledge reasoning, with typical trajectory horizons ranging from a few turns (Search-QA) to tens of turns (ALFWorld).

\subsection{Baselines}
\label{app:baselines}

We compare \method{} with a diverse set of baselines spanning no additional training, reinforcement learning, self-distillation, and teacher-based on-policy distillation. These baselines are chosen to reflect different supervision sources (environment rewards, self-generated targets, and external teacher distributions) and different signal granularities (trajectory-level rewards, token-level distillation, and temporally structured agent feedback).

\begin{itemize}[leftmargin=*]
  \item \textbf{Initial.} The base student model without any additional task-specific post-training. This baseline measures the zero-shot or instruction-tuned agentic capability of the underlying Qwen3 student before reinforcement learning or distillation.

  \item \textbf{GRPO (Group Relative Policy Optimization).} A reinforcement learning baseline that optimizes the student directly with environment rewards~\citep{shao2024deepseekmath}. For each input, GRPO samples a group of candidate trajectories, assigns each trajectory a scalar task reward, and normalizes rewards within the group to estimate relative advantages. It avoids training a separate value model, making it efficient for LLM post-training, but the reward is usually sparse and trajectory-level in multi-turn agent tasks, so all tokens in a trajectory receive relatively coarse credit.

  \item \textbf{SDAR (Self-Distilled Agentic Reinforcement Learning).} SDAR keeps RL as the primary optimization backbone and adds a gated self-distillation objective as an auxiliary signal~\citep{lu2026sdar}. Instead of relying on an external teacher, it uses the agent's own high-quality behaviors or self-generated targets to provide additional token-level guidance. This can improve credit assignment when the self-distilled signal is reliable, but its effectiveness depends on whether the student can already generate sufficiently useful agent trajectories.

  \item \textbf{OPD (On-Policy Distillation).} OPD trains the student on trajectories sampled from the student itself while using an external teacher distribution to provide dense token-level supervision~\citep{agarwal2024onpolicy,gu2024minillm}. Compared with offline supervised imitation, OPD reduces train--test mismatch because supervision is applied on the student's own rollout states. However, pure OPD mainly imitates the teacher and does not directly optimize environment rewards, so it may saturate once the student approaches teacher-like behavior.

  \item \textbf{SOD (Step-wise On-policy Distillation).} SOD is an agent-oriented OPD method that adaptively adjusts distillation strength at different interaction steps~\citep{zhong2026sod}. It is designed for long-horizon tool-use or agent trajectories, where different steps may have different levels of reliability and learning value. By assigning step-wise distillation weights, SOD provides more structured teacher guidance than uniform OPD, but it still relies primarily on teacher-driven supervision.

  \item \textbf{TCOD (Temporal Curriculum On-Policy Distillation).} TCOD introduces a temporal curriculum for multi-turn OPD~\citep{wang2026tcod}. Rather than exposing the student to the full trajectory difficulty uniformly, it controls the trajectory depth or temporal range used for distillation so that the student learns agent behavior in a progressively structured manner. This curriculum can stabilize multi-turn distillation, but its performance depends on the chosen temporal schedule.
\end{itemize}

The 4B GRPO model is reported as a teacher reference in the main table. All teacher-based distillation baselines use the same teacher checkpoint where applicable, so differences among OPD, SOD, TCOD, and \method{} mainly come from how teacher supervision is weighted and combined with reward optimization.

\subsection{Training Hyperparameters}
\label{app:config}

We conduct training across three text-based interactive environments: ALFWorld, Search-QA, and WebShop. The training and core \method{} hyperparameters are summarized in \cref{tab:hyperparams}, while task-specific configurations are provided in \cref{tab:hyperparams-task}.

\begin{table}[htbp]
\centering
\caption{Training and core \method{} hyperparameters.}
\label{tab:hyperparams}
\footnotesize
\setlength{\tabcolsep}{4.5pt}
\renewcommand{\arraystretch}{1.08}
\begin{tabular}{@{}ll@{}}
\toprule
\textbf{Hyperparameter} & \textbf{Value} \\
\midrule
\multicolumn{2}{c}{\textit{Training Setup}} \\
\midrule
Actor learning rate & $1\times10^{-6}$ \\
Actor KL loss & Off (\texttt{use\_kl\_loss=False}) \\
Group size (rollouts per prompt) & 8 \\
Max response length & 512 \\
Rollout engine & vLLM \\
GPU count & 8 (single node) \\
Training steps & 150 \\
Sampling temperature & 1.0 \\
Val temperature & 0.4 \\
Enable thinking & Off \\
\midrule
\multicolumn{2}{c}{\textit{Annealing Core Configuration}} \\
\midrule
KL coefficient initial $\beta_{\text{kl}}(0)$ & 1.0 \\
RL coefficient initial $\beta_{\text{rl}}(0)$ & 1.0 \\
Coefficient annealing & On \\
Annealing steps $T$ & 80 \\
KL coefficient minimum $\beta_{\text{kl}}^{\min}$ & 0.1 \\
RL coefficient maximum $\beta_{\text{rl}}^{\max}$ & 2.0 \\
Annealing schedule & Linear: $\beta_{\text{kl}}$ decays $1.0\rightarrow0.1$, $\beta_{\text{rl}}$ grows $1.0\rightarrow2.0$, clamped after $T$ \\
\bottomrule
\end{tabular}
\end{table}

\begin{table}[htbp]
\centering
\caption{Task-specific hyperparameters for ALFWorld, Search-QA, and WebShop.}
\label{tab:hyperparams-task}
\footnotesize
\setlength{\tabcolsep}{6pt}
\renewcommand{\arraystretch}{1.08}
\begin{tabular}{@{}lccc@{}}
\toprule
\textbf{Hyperparameter} & \textbf{ALFWorld} & \textbf{Search-QA} & \textbf{WebShop} \\
\midrule
Train batch size & 16 & 128 & 16 \\
Val batch size & 128 & 512 & 128 \\
Max prompt length & 2048 & 4096 & 4096 \\
Environment max steps & 50 & 4 & 15 \\
\bottomrule
\end{tabular}
\end{table}

\subsection{Annealing Parameter Ablation}
\label{app:annealing-ablation}

\begin{figure}[htbp]
\centering
\includegraphics[width=\textwidth]{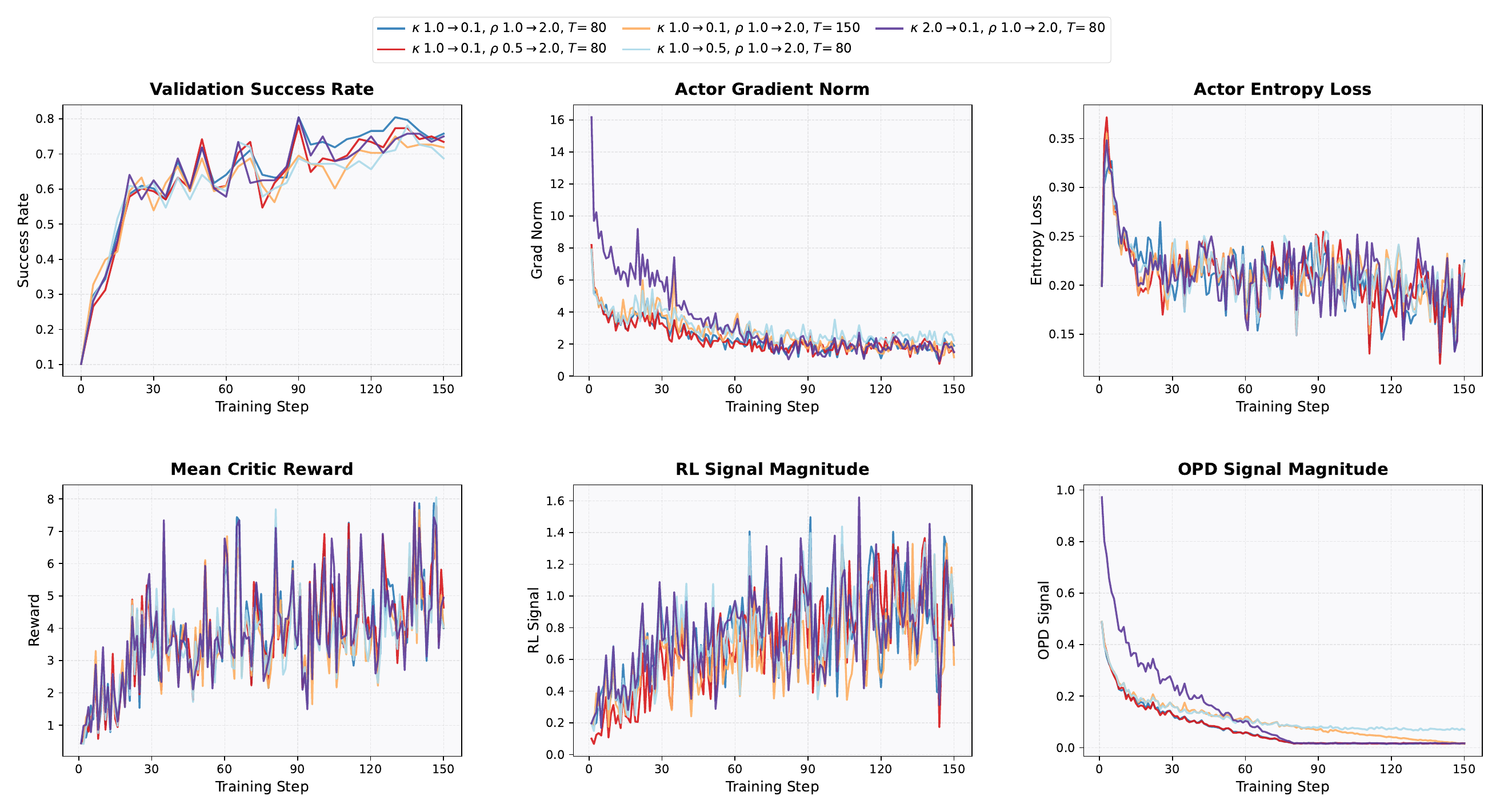}
\caption{Annealing-parameter ablation on ALFWorld over 150 training steps. Each curve corresponds to a different linear schedule for the OPD coefficient $\kappa$ and RL coefficient $\rho$, varying their initial or terminal values and the annealing horizon $T$. The six panels show validation success rate and the associated actor, critic, and hybrid-signal diagnostics.}
\label{fig:annealing-ablation}
\end{figure}

We conduct an annealing-parameter ablation on ALFWorld by varying the initialization, terminal values, and duration of the OPD/RL coefficient schedules. The complete settings and results are shown in \cref{fig:annealing-ablation}.

\subsection{Additional Training Diagnostics}
\label{app:additional-diagnostics}

\begin{figure}[htbp]
\centering
\includegraphics[width=\textwidth]{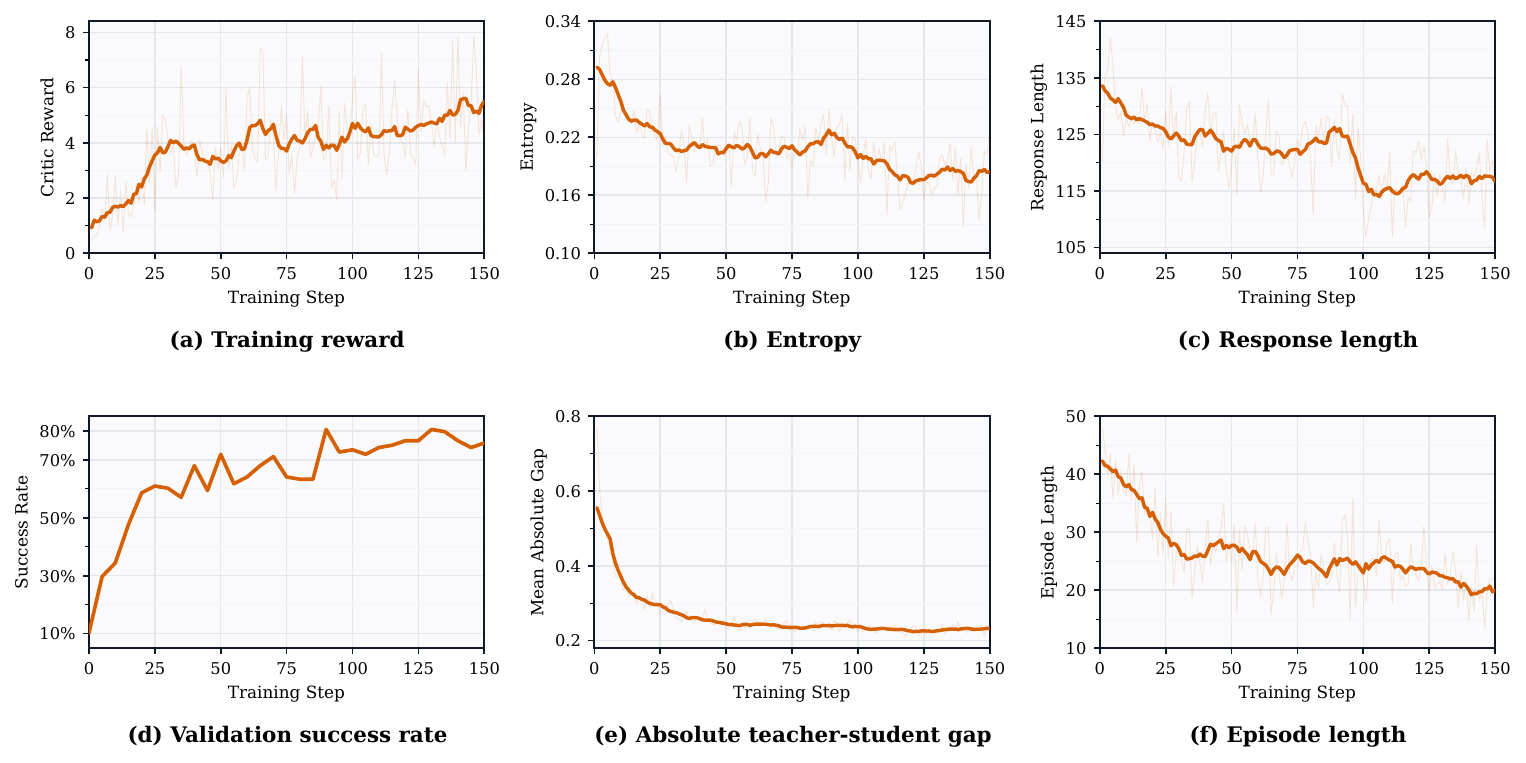}
\caption{Additional training diagnostic metrics for the 1.7B model on the ALFWorld dataset.}
\label{fig:atod-diagnostics}
\end{figure}

We report additional training diagnostics for the 1.7B model on ALFWorld, including training reward, entropy, response length, validation success rate, teacher--student gap, and episode length. The complete curves are shown in \cref{fig:atod-diagnostics}.

\subsection{T-DUR Diagnostic Metrics on ALFWorld}
\label{app:tdur-metrics}

\begin{figure}[htbp]
\centering
\includegraphics[width=\textwidth]{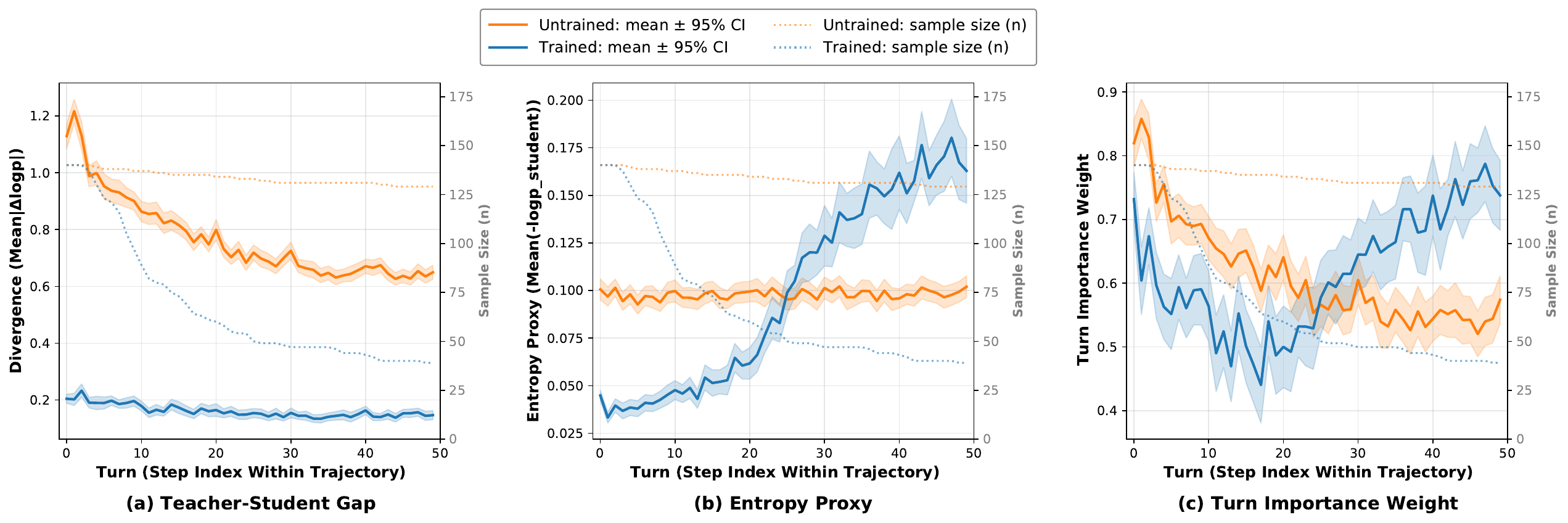}
\caption{T-DUR diagnostic metrics for the 1.7B student model on the ALFWorld validation set, comparing pre-trained and post-trained indicators. In all panels, the dashed line shows the number of valid evaluation trajectories (sample count) at each training step. \textbf{Panel (a):} Teacher--student gap (disagreement proxy $d_k$) over steps, showing how the divergence between teacher and student distributions shrinks as training progresses. \textbf{Panel (b):} Entropy proxy ($h_k$) distribution across training steps, reflecting the student's per-turn uncertainty before and after \method{} training. \textbf{Panel (c):} Turn-level T-DUR weight ($w_k$) as a function of training step, illustrating how the Soft-OR fusion dynamically reallocates relative distillation emphasis across turns. Together, these diagnostics show that T-DUR preserves larger relative weights on turns with high disagreement or uncertainty while the overall disagreement and entropy decrease with training.}
\label{fig:tdur-diagnostics}
\end{figure}

As shown in \cref{fig:tdur-diagnostics}, the dashed line in each panel indicates the number of evaluation trajectories available at each training step. We compare the diagnostic curves of the untrained (pre-training) and trained (post-training) 1.7B student to illustrate how \tdur{} allocates relative distillation weights.

\textbf{Untrained model.} Before training, the model achieves very low accuracy and exhibits an average trajectory length of 47.91 turns, which approaches the experimental maximum of 50 steps. This indicates that most samples produce long yet incorrect trajectories that exhaust the step budget. Panel (a) reveals a large teacher--student disagreement gap in the first few turns, where the teacher provides strong corrective signals against the student's erroneous actions. However, the gap progressively decays at later turns: as earlier mistakes accumulate into a long erroneous prefix, the teacher's distribution becomes less informative for distinguishing corrective actions, causing the disagreement to weaken. Panel (b) shows that the entropy proxy remains around 0.1 with little variation throughout the trajectory, reflecting that the untrained student stays in a persistently uncertain and error-prone regime across all turns. Consequently, Panel (c) shows that \tdur{} preserves larger relative weights on earlier turns and progressively attenuates later turns more strongly. This pattern is well-motivated: when the prompt prefix gradually drifts due to accumulated errors, the learning value of later turns diminishes, and the relative distillation emphasis should instead favor the earlier turning points and steps where mistakes first arise.

\textbf{Trained model.} After \method{} training, the teacher--student disagreement gap drops substantially compared to the untrained case (Panel a), confirming effective knowledge transfer. The entropy proxy (Panel b) exhibits a clear two-phase pattern: entropy remains at a low value during the first approximately 20 steps, then rises beyond step 20. This aligns with the observation that the trained model achieves markedly higher accuracy, with most successful trajectories completing within 25 turns; on these successful trajectories the student is confident and entropy is low. For trajectories that extend beyond 20 steps, early deviations or errors have likely already occurred, causing the student's uncertainty to increase at later positions. Correspondingly, Panel (c) shows that \tdur{} strongly attenuates the earlier, more confident turns while preserving larger relative weights on later turns where the student exhibits higher confusion. This asymmetric gating strategy prioritizes problematic decision points relative to routine turns the student has already mastered.

\section{Qualitative Examples and Diagnostic Analyses}
\label{app:qualitative-diagnostics}

This section provides concrete rollouts and diagnostic visualizations that explain where \tdur{} allocates teacher supervision in long trajectories.

\subsection{Environment Prompt Templates}
\label{app:prompt-templates}

Figures~\ref{fig:prompt-alfworld}--\ref{fig:prompt-webshop} present the full prompt templates used by \method{} for the three evaluation environments.

\begin{figure}[htbp]
\centering
\setlength{\fboxsep}{0pt}
\setlength{\fboxrule}{0.45pt}
\fbox{%
\begin{minipage}{0.97\textwidth}
\begingroup\setlength{\fboxsep}{4pt}\colorbox{casebar}{\parbox{\dimexpr\linewidth-8pt\relax}{\textcolor{white}{\textbf{Prompt of \method{} on ALFWorld}}}}\endgroup

\vspace{6pt}
\small\raggedright
\setlength{\leftskip}{8pt}
\setlength{\rightskip}{8pt plus 2em}
\setlength{\parindent}{0pt}
\setlength{\parskip}{3pt}
You are an expert agent operating in the ALFRED Embodied Environment. Your task is to: \texttt{\{task\_description\}}.

Prior to this step, you have already taken \texttt{\{step\_count\}} step(s). Below are the most recent \texttt{\{history\_length\}} observations and the corresponding actions you took: \texttt{\{action\_history\}}

You are now at step \texttt{\{current\_step\}} and your current observation is: \texttt{\{current\_observation\}}

Your admissible actions of the current situation are: \texttt{[[\{admissible\_actions\}]]}.

Now it's your turn to take an action. You should first reason step-by-step about the current situation. This reasoning process MUST be enclosed within \texttt{\textless think\textgreater{}} \texttt{\textless/think\textgreater{}} tags. Once you've finished your reasoning, you should choose an admissible action for current step and present it within \texttt{\textless action\textgreater{}} \texttt{\textless/action\textgreater{}} tags.
\vspace{4pt}
\end{minipage}}
\caption{Prompt template used by \method{} for the ALFWorld task environment.}
\label{fig:prompt-alfworld}
\end{figure}

\begin{figure}[htbp]
\centering
\setlength{\fboxsep}{0pt}
\setlength{\fboxrule}{0.45pt}
\fbox{%
\begin{minipage}{0.97\textwidth}
\begingroup\setlength{\fboxsep}{4pt}\colorbox{casebar}{\parbox{\dimexpr\linewidth-8pt\relax}{\textcolor{white}{\textbf{Prompt of \method{} on Search-based QA}}}}\endgroup

\vspace{6pt}
\small\raggedright
\setlength{\leftskip}{8pt}
\setlength{\rightskip}{8pt plus 2em}
\setlength{\parindent}{0pt}
\setlength{\parskip}{3pt}
You are an expert agent tasked with answering the given question step-by-step.

Your question: \texttt{\{task\_description\}}.

Prior to this step, you have already taken \texttt{\{step\_count\}} step(s). Below is the interaction history where \texttt{\textless search\textgreater{}} \texttt{\textless/search\textgreater{}} wrapped your past search queries and \texttt{\textless information\textgreater{}} \texttt{\textless/information\textgreater{}} wrapped the corresponding search results returned by the external search engine. History:

\texttt{\{memory\_context\}}

Now it's your turn to respond for the current step. You should first conduct a reasoning process. This process MUST be enclosed within \texttt{\textless think\textgreater{}} \texttt{\textless/think\textgreater{}} tags. After completing your reasoning, choose only one of the following actions (do not perform both):
\begin{enumerate}[leftmargin=18pt,labelsep=6pt,itemsep=1pt,topsep=2pt]
  \item If you find you lack some knowledge, you MUST call a search engine to get more external information using format: \texttt{\textless search\textgreater{}} your query \texttt{\textless/search\textgreater{}}.
  \item If you have enough knowledge to answer the question confidently, provide your final answer within \texttt{\textless answer\textgreater{}} \texttt{\textless/answer\textgreater{}} tags, without detailed illustrations. For example, \texttt{\textless answer\textgreater{}Beijing\textless/answer\textgreater{}}.
\end{enumerate}
\vspace{4pt}
\end{minipage}}
\caption{Prompt template used by \method{} for the Search-based QA task environment.}
\label{fig:prompt-searchqa}
\end{figure}

\begin{figure}[htbp]
\centering
\setlength{\fboxsep}{0pt}
\setlength{\fboxrule}{0.45pt}
\fbox{%
\begin{minipage}{0.97\textwidth}
\begingroup\setlength{\fboxsep}{4pt}\colorbox{casebar}{\parbox{\dimexpr\linewidth-8pt\relax}{\textcolor{white}{\textbf{Prompt of \method{} on WebShop}}}}\endgroup

\vspace{6pt}
\small\raggedright
\setlength{\leftskip}{8pt}
\setlength{\rightskip}{8pt plus 2em}
\setlength{\parindent}{0pt}
\setlength{\parskip}{3pt}
You are an expert autonomous agent operating in the WebShop e-commerce environment.

Your task is to: \texttt{\{task\_description\}}.

Prior to this step, you have already taken \texttt{\{step\_count\}} step(s). Below are the most recent \texttt{\{history\_length\}} observations and the corresponding actions you took: \texttt{\{action\_history\}}

You are now at step \texttt{\{current\_step\}} and your current observation is: \texttt{\{current\_observation\}}.

Your admissible actions of the current situation are: \texttt{[\{available\_actions\}]}.

Now it's your turn to take one action for the current step. You should first reason step-by-step about the current situation, then think carefully which admissible action best advances the shopping goal. This reasoning process MUST be enclosed within \texttt{\textless think\textgreater{}} \texttt{\textless/think\textgreater{}} tags. Once you've finished your reasoning, you should choose an admissible action for current step and present it within \texttt{\textless action\textgreater{}} \texttt{\textless/action\textgreater{}} tags.
\vspace{4pt}
\end{minipage}}
\caption{Prompt template used by \method{} for the WebShop task environment.}
\label{fig:prompt-webshop}
\end{figure}

\subsection{Turn-level Case Studies}
\label{app:case-studies}

We include two compact case studies in \cref{fig:case-study-alfworld,fig:case-study-webshop} to visualize how \tdur{} assigns turn-level distillation weights. In each panel, $d_k$ and $h_k$ are the raw, pre-normalization proxies for teacher--student disagreement and sampled-token student uncertainty, respectively, while $w_k$ is the final Soft-OR turn weight after per-trajectory normalization. The ALFWorld trace is from a trained student; the WebShop trace uses a pre-distillation student so that the reasoning-rich error-and-recovery pattern remains visible, with metrics computed against the same teacher. The layout follows the qualitative diagnostic style in which the task context appears above a step-wise trace, and the most important weight column is highlighted.

\begin{figure}[htbp]
\centering
\setlength{\fboxsep}{0pt}
\setlength{\fboxrule}{0.45pt}
\fbox{%
\begin{minipage}{0.97\textwidth}
\begingroup\setlength{\fboxsep}{4pt}\colorbox{casebar}{\parbox{\dimexpr\linewidth-8pt\relax}{\textcolor{white}{\textbf{Case A: ALFWorld --- T-DUR Highlights Planning, Route Choice, and Goal Completion}}}}\endgroup

\vspace{5pt}
\begingroup\setlength{\fboxsep}{5pt}\colorbox{gray!8}{\begin{minipage}{0.965\linewidth}
\textbf{Task:} clean some lettuce and put it in fridge.\hfill
\textbf{Outcome:} success, 7 turns.\\
\textbf{Reading guide:} $d_k$ is raw teacher--student disagreement, $h_k$ is raw sampled-token student uncertainty, and $w_k$ is the final Soft-OR weight after normalizing $d_k$ and $h_k$ within the trajectory.
\end{minipage}}\endgroup

\vspace{4pt}
\scriptsize
\setlength{\tabcolsep}{3pt}
\renewcommand{\arraystretch}{1.08}
\begin{tabular}{@{}C{0.055\linewidth}L{0.62\linewidth}C{0.075\linewidth}C{0.075\linewidth}C{0.075\linewidth}@{}}
\toprule
\textbf{Step} & \textbf{Student reasoning trace (abbreviated)} & $\boldsymbol{d_k}$ & $\boldsymbol{h_k}$ & $\boldsymbol{w_k}$ \\
\midrule
$s_0$ & \textbf{Obs.} The room contains countertops, cabinets, a fridge, and a sinkbasin. \textbf{Student.} ``I first need to find the lettuce; I will start by checking countertops.'' \textbf{Action:} \texttt{go to countertop 1}. \emph{Open-ended planning has the trajectory's highest uncertainty.} & 0.101 & 0.037 & \casew{casegreen}{1.00} \\
\midrule
$s_1$ & \textbf{Obs.} Countertop 1 contains \texttt{lettuce 1}. \textbf{Student.} ``I should take the lettuce first; after cleaning, I can put it in the fridge.'' \textbf{Action:} \texttt{take lettuce 1 from countertop 1}. & 0.120 & 0.013 & \casew{caseyellow}{0.60} \\
\midrule
$s_2$ & \textbf{Obs.} The lettuce is now in inventory. \textbf{Student.} ``The next logical step is to clean it at sinkbasin 1.'' \textbf{Action:} \texttt{go to sinkbasin 1}. \emph{This route-choice turn has the largest disagreement signal.} & 0.228 & 0.025 & \casew{casegreen}{1.00} \\
\midrule
$s_3$ & \textbf{Obs.} At sinkbasin 1; admissible actions include cleaning the lettuce. \textbf{Student.} ``The next step is to clean the lettuce using the sinkbasin.'' \textbf{Action:} \texttt{clean lettuce 1 with sinkbasin 1}. \emph{A mastered mechanical step.} & 0.022 & 0.010 & \casew{gray!15}{0.15} \\
\midrule
$s_4$ & \textbf{Obs.} The lettuce has been cleaned. \textbf{Student.} ``Now I need to put it in the fridge; go to fridge 1.'' \textbf{Action:} \texttt{go to fridge 1}. & 0.054 & 0.022 & \casew{caseyellow}{0.60} \\
\midrule
$s_5$ & \textbf{Obs.} At fridge 1; the fridge is closed. \textbf{Student.} ``I need to open the fridge.'' \textbf{Action:} \texttt{open fridge 1}. \emph{Lowest on both signals, so distillation can skip it.} & 0.019 & 0.006 & \casew{gray!15}{0.00} \\
\midrule
$s_6$ & \textbf{Obs.} The fridge is open. \textbf{Student.} ``Place the cleaned lettuce into the fridge.'' \textbf{Action:} \texttt{move lettuce 1 to fridge 1}. \emph{The goal-completing action is recovered by disagreement despite low entropy.} & 0.138 & 0.008 & \casew{caseyellow}{0.60} \\
\bottomrule
\end{tabular}
\vspace{3pt}
\end{minipage}}
\caption{ALFWorld appendix case study. \tdur{} does not simply favor long or late turns: it preserves larger relative weights on the initial search plan, the route decision toward cleaning, and the final placement, while assigning near-zero weight to routine environment mechanics such as opening the fridge.}
\label{fig:case-study-alfworld}
\end{figure}

\begin{figure}[htbp]
\centering
\setlength{\fboxsep}{0pt}
\setlength{\fboxrule}{0.45pt}
\fbox{%
\begin{minipage}{0.97\textwidth}
\begingroup\setlength{\fboxsep}{4pt}\colorbox{casebar}{\parbox{\dimexpr\linewidth-8pt\relax}{\textcolor{white}{\textbf{Case B: WebShop --- T-DUR Flags Error-and-Recovery Turns in Product Selection}}}}\endgroup

\vspace{5pt}
\begingroup\setlength{\fboxsep}{5pt}\colorbox{gray!8}{\begin{minipage}{0.965\linewidth}
\textbf{Task:} find a machine-wash men's dress shirt with cotton spandex, classic fit, short sleeve, shipwreck burgundy, size 2x, and price lower than \$60.\\
\textbf{Reading guide:} $d_k$ is raw teacher--student disagreement, $h_k$ is raw sampled-token student uncertainty, and $w_k$ is the final Soft-OR weight after normalizing $d_k$ and $h_k$ within the trajectory.
\end{minipage}}\endgroup

\vspace{4pt}
\scriptsize
\setlength{\tabcolsep}{3pt}
\renewcommand{\arraystretch}{1.08}
\begin{tabular}{@{}C{0.055\linewidth}L{0.62\linewidth}C{0.075\linewidth}C{0.075\linewidth}C{0.075\linewidth}@{}}
\toprule
\textbf{Step} & \textbf{Student reasoning trace (abbreviated)} & $\boldsymbol{d_k}$ & $\boldsymbol{h_k}$ & $\boldsymbol{w_k}$ \\
\midrule
$s_0$ & \textbf{Obs.} The page is at search. \textbf{Student.} ``Search for the exact shirt constraints: cotton spandex, classic fit, short sleeve, shipwreck burgundy, size 2x, under \$60.'' \textbf{Action:} \texttt{search[... size 2x, under \$60]}. & 0.796 & 0.063 & \casew{gray!15}{0.23} \\
\midrule
$s_1$ & \textbf{Obs.} Search results show many candidate products. \textbf{Student.} The model opens a seemingly relevant item for inspection. \textbf{Action:} \texttt{click[b07hrfsnl4]}. \emph{Candidate selection raises uncertainty.} & 0.705 & 0.132 & \casew{caseyellow}{0.48} \\
\midrule
$s_2$ & \textbf{Obs.} The item page exposes size and color options. \textbf{Student.} The model tries to adjust size but clicks the wrong option. \textbf{Action:} \texttt{click[x-small]}. \emph{Both uncertainty and disagreement fire on the error turn.} & 0.913 & 0.195 & \casew{casegreen}{0.92} \\
\midrule
$s_3$ & \textbf{Obs.} Same item page with the wrong size active. \textbf{Student.} ``The correct size is 2x; click 2x.'' \textbf{Action:} \texttt{click[2x]}. \emph{A high-value recovery turn.} & 0.876 & 0.158 & \casew{caseorange}{0.76} \\
\midrule
$s_4$ & \textbf{Obs.} Same item page. \textbf{Student.} The model re-clicks the already intended size. \textbf{Action:} \texttt{click[2x]}. \emph{A consolidation step with lower utility.} & 0.668 & 0.101 & \casew{gray!15}{0.25} \\
\midrule
$s_5$ & \textbf{Obs.} Color options are available. \textbf{Student.} ``Select the required shipwreck burgundy color.'' \textbf{Action:} \texttt{click[shipwreck burgundy]}. & 0.696 & 0.132 & \casew{caseyellow}{0.47} \\
\midrule
$s_6$ & \textbf{Obs.} The filters are not visibly narrowing the result. \textbf{Student.} The model keeps adjusting size/color constraints. \textbf{Action:} \texttt{click[2x]}. \emph{This is the trajectory peak for both disagreement and uncertainty.} & 1.214 & 0.219 & \casew{casegreen}{1.00} \\
\midrule
$s_7$ & \textbf{Obs.} The item page has \texttt{buy now}. \textbf{Student.} ``The selected size and color satisfy the request; proceed with purchase.'' \textbf{Action:} \texttt{click[buy now]}. \emph{The commit step remains important.} & 0.744 & 0.158 & \casew{caseorange}{0.67} \\
\bottomrule
\end{tabular}
\vspace{3pt}
\end{minipage}}
\caption{WebShop appendix case study. Unlike the ALFWorld example, WebShop contains free-text search and many clickable attributes, so disagreement remains high even when the student appears decisive. The Soft-OR weight therefore captures both hesitant turns and confidently different turns such as the wrong-size click and the final purchase commitment.}
\label{fig:case-study-webshop}
\end{figure}

\end{document}